\documentclass{article}

\usepackage{microtype}
\usepackage{graphicx}
\usepackage{subfigure}
\usepackage{booktabs} % for professional tables

\usepackage{hyperref}

\usepackage[accepted]{icml2024}

\usepackage{amsmath}
\usepackage{amssymb}
\usepackage{mathtools}
\usepackage{amsthm}

\usepackage[capitalize,noabbrev]{cleveref}

\theoremstyle{plain}
\newtheorem{theorem}{Theorem}[section]

\theoremstyle{definition}
\newtheorem{definition}[theorem]{Definition}

\theoremstyle{remark}

\usepackage[textsize=tiny]{todonotes}

\usepackage{url}            % simple URL typesetting
\usepackage{algorithm}
\usepackage{algorithmic}
\usepackage{esvect}
\usepackage{color, colortbl}
\definecolor{Gray}{gray}{0.9}
\usepackage{xcolor}
\usepackage[normalem]{ulem}
\renewcommand{\algorithmicrequire}{\textbf{Input:}}
\renewcommand{\algorithmicensure}{\textbf{Output:}}
\usepackage{subcaption}

\usepackage{mathtools}
\usepackage{amsthm}
\usepackage{float}
\usepackage{multirow}
\usepackage{amsmath}
\usepackage{amsthm}
\usepackage{amssymb}
\usepackage{graphicx}
\usepackage{xargs}[2008/03/08]
\PassOptionsToPackage{normalem}{ulem}
\usepackage{ulem}
\usepackage{wrapfig}

\DeclareMathOperator*{\argmax}{arg\,max}
\DeclareMathOperator*{\argmin}{arg\,min}

\icmltitlerunning{Controllable Prompt Tuning For Balancing Group Distributional Robustness}

\begin{document}

\twocolumn[
\icmltitle{Controllable Prompt Tuning For Balancing Group Distributional Robustness}



\icmlsetsymbol{equal}{*}

\begin{icmlauthorlist}
\icmlauthor{Hoang Phan}{nyu}
\icmlauthor{Andrew Gordon Wilson}{nyu}
\icmlauthor{Qi Lei}{nyu}
\end{icmlauthorlist}

\icmlaffiliation{nyu}{New York University}
\icmlcorrespondingauthor{Hoang Phan}{hvp2011@nyu.edu}

\icmlkeywords{Machine Learning, ICML}

\vskip 0.3in
]



\printAffiliationsAndNotice{}  % leave blank if no need to mention equal contribution

\begin{abstract}
%In this paper, we study the problem of spurious correlation with the availability of group annotated training data. Existing works in this direction often adopt heuristic strategies to encourage the model to fit the smaller groups by subsampling majority groups or upweighting minority ones. 
%However, our investigation shows that they are not efficient in optimizing the performance across groups yet focus on optimizing the worst group only. 

%Motivated by the multiple-objective nature of the optimization problem for group robustness, we propose a principled algorithm to balance the performance trade-off between subgroups, 
%conditioning on a pre-defined vector. While empirical results verify the effectiveness of our algorithm, directly applying such optimization to large-scale problems involves updating the parameters of the entire network, making it both computationally expensive and challenging. Thus, we introduce Controllable Prompt Tuning (CPT) that couples the introduced algorithm with prompt-tuning techniques in order to make our method scalable when the number of controlling vectors grows.  Finally, we conduct extensive experiments on both transformer and non-transformer architectures as well as on unimodal and multimodal settings, in which our method surpasses prior methods to achieve state-of-the-art with only $0.4\%$ tuned parameters. Our implementation is provided in \url{https://anonymous.4open.science/r/CPT/}.

Models trained on data composed of different groups or domains can suffer from severe performance degradation under distribution shifts. While recent methods have largely focused on optimizing the worst-group objective, this often comes at the expense of good performance on other groups. To address this problem, we introduce an optimization scheme to achieve good performance across groups and find a good solution for all without severely sacrificing performance on any of them. However, directly applying such optimization involves updating the parameters of the entire network, making it both computationally expensive and challenging. Thus, we introduce Controllable Prompt Tuning (CPT), which couples our approach with prompt-tuning techniques. On spurious correlation benchmarks, our procedures achieve state-of-the-art results across both transformer and non-transformer architectures, as well as unimodal and multimodal data, while requiring only $0.4\%$ tunable parameters.

%Optimizing machine learning problems with multiple groups/tasks suffers from performance degradation due to distributional shifts. In this work, we propose a principled optimization scheme to balance the performance trade-off between groups with multi-objective optimization (MOO) strategies. 
%We design individual objective functions that incorporate the effect of heterogeneity among tasks (different sample sizes and inconsistent convergence speeds) and integrate them with the MOO component that intends to find their Pareto optimality. To further enhance training efficiency and scalability, we introduce Controllable Prompt Tuning (CPT) which couples the introduced algorithm with prompt-tuning techniques. Finally, we conduct extensive experiments on both transformer and non-transformer architectures as well as on unimodal and multimodal settings, in which our method surpassed prior methods and achieved state-of-the-art with only $0.4\%$ tuned parameters. Our implementation is available in \url{https://anonymous.4open.science/r/CPT/}.
\end{abstract}

% \vspace*{-1mm}
\section{Introduction}
\label{sec:intro}
% \vspace*{-1mm}

Spurious correlation or shortcut learning arises when a classifier relies on non-predictive features that coincidentally correlate with class labels among training samples.  For example, a \emph{majority} of waterbirds are often found near water, while landbirds mostly appear near land. Even when being provided with a small number of \emph{minority group examples}, such as waterbirds on land, or landbirds on water, model predictions are often still dominated by spurious features, in this case, the background, ignoring the salient feature, the foreground \citep{sagawa2019distributionally}.

A variety of approaches aim to improve worst group performance, by oversampling high-loss samples~\citep{liu2021just, zhang2022correct}, undersampling the majority group \citep{sagawa2020investigation} or re-training a last layer on group-balanced validation data \citep{kirichenko2022last}. A particularly popular and effective baseline, \emph{Group Distributionally Robust Optimization} (GroupDRO) \citep{sagawa2019distributionally, sagawa2020investigation}, directly minimizes an estimated upper bound on worst group loss. However, all these procedures generally neglect knowledge transfer between groups. Moreover, high-error samples can potentially include noisy-label data that could hurt the model's predictive ability if we try to learn them exhaustively \citep{oh2022improving}. %\agw{awkward phrasing (``learn them exhaustively'')}. 
Besides, DRO-based methods in particular are susceptible to overfitting, wherein their test performances can decline to the same as ERM with sufficiently long training \citep{gulrajani2020search, piratla2021focus, zhai2022understanding}.

%they can also lead to overfitting to a particular group, especially if there are noisy labels \citep{oh2022improving}.

%
%Various approaches discourage the model from learning spurious features by oversampling high-loss samples ~\citep{liu2021just, zhang2022correct}, minimizing the worst-case empirical risk \citep{sagawa2020investigation, piratla2021focus} or relying on auxiliary concept bank to de-bias the neural network \citep{wu2023discover, yang2023mitigating}.
%Among them, Group Distributionally Robust Optimization (GroupDRO) \citep{sagawa2019distributionally} is a well-known method that is often employed as an upper bound for other works in this line of research. However, the updating formula of GroupDRO directly minimizes an estimated upper bound on the worst group test loss, which unintentionally neglects the knowledge transfer between groups. Even worse, high-error samples can potentially include noisy-label data that could hurt the model's predictive ability if we try to learn them exhaustively \citep{oh2022improving}. For that reason, optimizing over the worst group might yield unsatisfactory results, i.e. when the model struggles to improve the performance of some specific group without considering the interplay of conflicts or cooperation between them. 

% \vspace*{-2mm}
\begin{figure}[!ht]
    \centering
     \includegraphics[width=1\columnwidth, ]{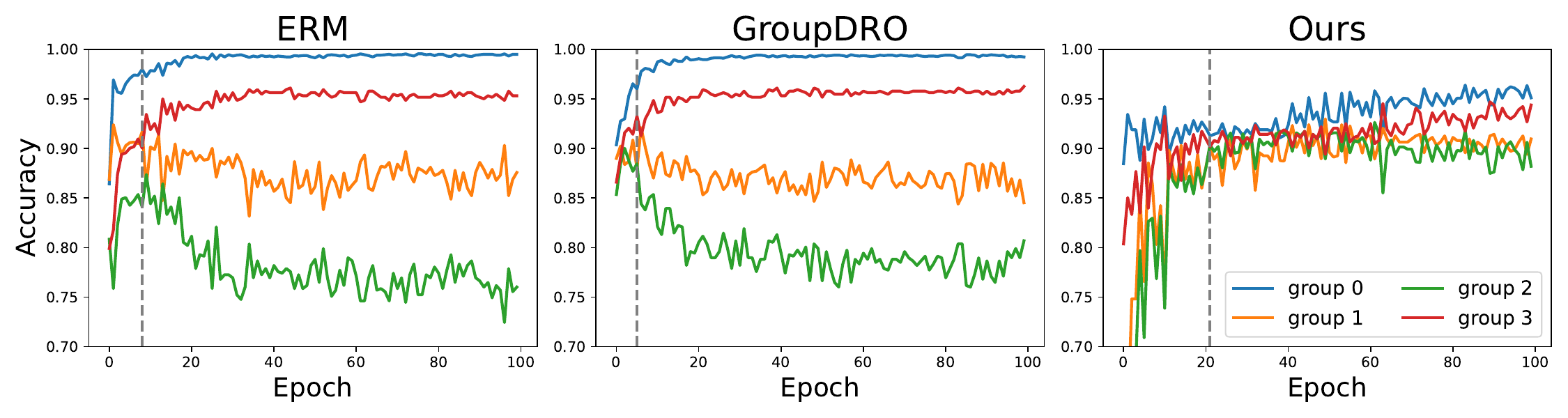}
     % \vspace*{-6mm}
    \caption{ Accuracy curves on Waterbirds for four groups during training. Vertical lines indicate early stopping epochs as models obtain the best performance on the validation set.}
    
    \label{fig:acc_curve}
\end{figure}
    % \vspace*{-2mm}

To visually illustrate these limitations, we plot in Figure \ref{fig:acc_curve} the performance of a ResNet50 on the Waterbirds benchmark, trained with ERM, GroupDRO, and our approach (to be explained later).
The plots demonstrate that while GroupDRO can improve the performance of the minority group (green line) early in training compared to ERM, it rapidly overfits and fails to maintain this performance over time. % later in training. 
After training for ten epochs, the test performance gap between the minority and majority (blue line) groups grows sharply. 
To address these critical limitations, we introduce a systematic approach to train a model that exhibits consistently high performance across all groups. Specifically, at each iteration, we identify a descending direction that benefits the groups simultaneously rather than focusing on an individual group. Figure \ref{fig:acc_curve} shows that our proposed method performs almost equally well in every group, thanks to our balancing mechanism.

Moreover, we aim to control the group-wise learning process by ensuring that the magnitude of loss for different groups is inversely proportional to a predefined vector $\mathbf{c}$. Varying this hyperparameter $\mathbf{c}$ adjusts the trade-off between the worst group and the average performance, yielding a more flexible method than prior work that only focuses on maximizing the worst group accuracy. However, computation can grow linearly with the number of controlling vectors $\mathbf{c}$. To overcome this challenge, we propose decoupling our optimization framework with parameter-efficient fine-tuning (PEFT) techniques that considerably reduce the number of trainable parameters. With the model compression ability powered by prompt-tuning, our proposed model can scale well for a large number of controlling vectors with a small parameter increase, e.g., $0.01\%$ and $0.003\%$ per value of $\mathbf{c}$ for Vision Transformer \citep{dosovitskiy2020image} and CLIP \citep{radford2021learning}, respectively.

\textbf{Contribution}. In this work, we introduce \textbf{CPT}, a \textbf{C}ontrollable \textbf{P}rompt \textbf{T}uning method to prevent models from learning spurious features. In summary, our key contributions are depicted as follows:

$\bullet$ Drawing inspiration from the principles of multi-objective optimization theory, we introduce a novel balancing mechanism to address the challenges of group distributional robustness. Our method not only considers the worst group but also leverages the gradient information from all groups to determine an ultimate updating direction that benefits all of them.

    $\bullet$ While the above rigid balancing procedure works well in different scenarios, we extend our method by introducing a controlling vector that allows us to dynamically adjust the priority across groups. Additionally, we integrate prompt tuning techniques to enhance the scalability of our method as the complexity and number of controlling vectors increase and to make it applicable for de-biasing large transformers-based models.
  
   $\bullet$
   We conduct intensive experiments on many different benchmarks, where CPT consistently exhibits superior performance. Our method surpasses the current state-of-the-art baselines on Waterbirds andimprove the performance  CelebA datasets while updating $0.4\%$ parameters. Moreover, it also outperforms recent proposed methods that aims to de-bias Vision Transformer and CLIP models with minimal training cost.

%%%\vspace*{-1mm}
\section{Related work}
\label{sec:related}
%%%\vspace*{-1mm}

We first review relevant methods to enhance distributional robustness and then discuss prior work related to the two essential parts of our proposed method: transformer and parameter-efficient fine-tuning. 

\paragraph{Distributional robustness}can be compromised for reasons like selection bias or spurious correlation across tasks. For selection bias on individual samples, prior works seek to reduce bias via practices like importance reweighting~\citep{heckman1979sample,shimodaira2000improving,cortes2010learning,lei2021near}, hard sample reweighting~\citep{liu2021just,nam2020learning} or distribution matching and discrepancy minimization~\citep{cortes2015adaptation,ben2010theory,berthelot2021adamatch}, and domain-adversarial algorithms~\citep{ganin2016domain,long2018conditional,phan2023global} across tasks in their feature representation space \citep{tran2023sharpness}. For selection bias on groups or subpopulation, namely subpopulation shift or dataset imbalance, label propagation ~\citep{cai2021theory,berthelot2021adamatch} or other consistency regularization~\citep{miyato2018virtual,yang2023sample}  
 are used to generalize the prediction to broader domains. %In particular, GroupDRO~\citep{sagawa2019distributionally, sagawa2020investigation} or other relevant methods~\citep{krueger2021out,hu2018does} select descending direction more in favor of the worse performing/minority group(s) during training. 
 
Improving distributional robustness also requires mitigating the effect of environmental features. Prior work uses model ensemble~\citep{kumar2022calibrated} or model soups~\citep{wortsman2022model} to learn rich features and reduce the effect of spurious features.  More practices include invariant feature representation learning~\citep{arjovsky2019invariant,chen2022iterative}, domain generalization \citep{krueger2021out, sun2017correlation, shi2021gradient} or through data augmentation \citep{xu2020adversarial,yao2022improving}.
 
%\textbf{Distributional Robust Optimization}: %Recent works aim to build deep learning models that are robust against spurious correlations, .  Among those methods that utilize group labels, GroupDRO \citep{sagawa2019distributionally}  is shown to be effective via minimizing the error on the group with the highest loss during training.

\textbf{Transformer.} In the past few years, transformer \citep{vaswani2017attention} has emerged as a dominant architecture for a wide range of applications, exhibiting their profound capacity in modeling different modalities: languages \citep{devlin2018bert, liu2019roberta,  touvron2023llama}, vision \citep{dosovitskiy2020image, chen2021crossvit} and speech \citep{radford2023robust, wang2023neural}. Due to its ability to process multiple modalities, transformer-based architectures were widely adopted for image understanding using different types of supervision. For example, Vision Transformer (ViT) \citep{dosovitskiy2020image} utilizes masked patches for pre-training and image-label supervision for fine-tuning.  Notably, Contrastive Language-Image Pre-training (CLIP) \citep{radford2021learning}, which was pre-trained on billion-scale image-text pairs, has demonstrated excellent zero-shot performance in image classification tasks. Despite obtaining impressive results on various downstream tasks, over-parameterization is still a crucial problem for transformer \citep{fan2019reducing, panahi2021shapeshifter} which potentially causes overfitting \citep{li2023dropkey} and unintended biases \citep{agarwal2021evaluating, wang2021gender, du2022learning, zhang2022contrastive} more severely than small-scale architectures \cite{sagawa2020investigation}.

\textbf{Parameter-Efficient Fine-Tuning.} While conventional training updates the entire network, prior work has shown that it is possible to achieve performance comparable with full fine-tuning by instead updating a small set of parameters \citep{lialin2023scaling, he2023sensitivity, he2021towards}. Early work in this line of research includes adapter \citep{houlsby2019parameter, chen2022adaptformer, sung2022vl}, which adds a lightweight fully connected network between the layers of a frozen pre-trained model. Since then, there have been many other methods that update subsets of the network \citep{guo2021parameter, gheini2021cross, zaken2022bitfit} or reparametrize the weights of the network using low-rank decompositions \citep{aghajanyan2020intrinsic, hu2021lora, edalati2022krona}. Another appealing PEFT strategy is prompt tuning, which prepends a few trainable tokens before the input sequence or hidden state to encode task-specific knowledge to the pre-trained model \citep{lester2021power, li2021prefix, zhu2023visual, huang2023diversity}.

\textbf{Gradient-based multi-task learning.} Prior work often searches for an update direction that benefits all tasks \cite{yu2020gradient, liu2021conflict, phan2022improving} or controls the trade-offs between per-task performance \cite{lin2019pareto, mahapatra2020multi, phan2022stochastic}. Therefore, casting the problem of learning from multiple domains into a multi-task problem \citep{liang2021pareto, kim2023removing} allows us to apply results from multi-objective optimization theory. Building on these insights, we propose a method that prevents the dominance of any single domain.

\vspace*{-1mm}
\section{Background}
\label{sec:background}
% \vspace*{-1mm}

In this section, we first formulate the problem of group robustness and recap the technical details of GroupDRO, then revisit some concepts of multi-objective optimization.
% \vspace*{-1mm}
\subsection{Problem formulation}
\vspace*{-1mm}

Formally, we consider the conventional setting of classifying an input $x \in \mathcal{X}$ as a target $y \in \mathcal{Y} $, where $\mathcal{X}, \mathcal{Y}$ are input and label spaces, respectively. We are given a training dataset composed of $K$ groups from the set $\mathcal{G}$, where each group $g \in \mathcal{G}$ consists of $n_g$ instances sampled from the probability distribution $P_g(\mathcal{X}, \mathcal{Y})$. Since the numbers of examples are different among groups in the training data, we consider groups with relatively large $n_g$ as majority groups and those with small $n_g$ as minority groups. Our main goal is to develop a model that performs effectively across all groups within $\mathcal{G}$  to avoid learning spurious correlations. 

Following previous methods \cite{sagawa2020investigation}, we adopt two metrics: worst group accuracy, indicating the minimum test accuracy across all groups $g \in \mathcal{G}$, and average accuracy, which represents the weighted average test accuracy with the weights corresponding to the relative proportions of each group in the training data. It is worth noting that commonly used datasets for group robustness often exhibit skewed training set distributions, the weighted average score is thus dominated by the performance of the model on groups that include large numbers of training data. Therefore, we propose to evaluate the model performance by additionally reporting the mean (unweighted average) accuracy score across different groups.
%\vspace*{-1mm}
\subsection{Group distributionally robust optimization}
%\vspace*{-1mm}

GroupDRO learns the model parameters $\theta$ by directly optimizing for the worst-group training loss as follows: %\vspace*{-2mm}
$$\max _{g \in \mathcal{G}} \mathbb{E}_{(x, y) \sim P_g(\mathcal{X}, \mathcal{Y}) }\left[\ell\left(f_\theta(x), y\right)\right] %\vspace*{-2mm}
$$
for some loss function $\ell: \mathcal{Y} \times \mathcal{Y} \rightarrow \mathbb{R}^+$. Then, their updating formula at each step is given by: 

$\quad g^*=\arg \max \ell_g(\theta) \quad$ and $\quad \theta^{t+1}=\theta^t-\eta \nabla_\theta \ell_{g^*}$, \;\;\;\; where $\eta$ is the learning rate.
%\vspace*{-1mm}
\subsection{Multi-objective Optimization}
%\vspace*{-1mm}

Assume that we are given $m$ objectives functions $f_{i}(x)$, $i\in[m]$
where $x\in\ensuremath{\mathbb{R}^{d}}$ and each $f_{i}:\ensuremath{\mathbb{R}^{d}}\ensuremath{\rightarrow\mathbb{R}}$
is a scalar-valued function (we use the $[m]$ notation to denote
the set $\{1,2,\dots,m\}$) . In the multi-objective optimization (MOO)
problem, we are interested in minimizing a vector-valued objective function
whose $i$-th component is $f_{i}(x)$:
\begin{equation}
\vec{F}(x)=[f_{1}(x),f_{2}(x),\dots,f_{m}(x)]\label{eq:moo} %\vspace*{-1mm}
\end{equation}

In general, an optimal solution for this objective vector function
does not exist since each of the individual function $f_{i}(x)$ does
not necessarily guarantee to have the same minimum solution. Alternatively,
we expect to obtain a solution, from which we cannot improve any specific
objective without hurting another. According to the above formulation, the solution \citep{zitzler1999multiobjective} of the multi-objective minimization problem is formally defined as follows:

\begin{definition}(Pareto dominance) Let $x_{1},x_{2}$ be two solutions for
the multi-objective optimization problem in Equation \ref{eq:moo},
$x_{1}$ dominates $x_{2}$ $(x_{1}\prec x_{2})$ if and only if $f_{i}(x_{1})\leq f_{i}(x_{2})\forall i\in[m]$
and $\exists j\in[m]$ s.t. $f_{j}(x_{1})<f_{j}(x_{2})$.
\end{definition}

\begin{definition}(Pareto optimality) A solution\textbf{ $x^{*}$ }in problem
\ref{eq:moo} is said to be Pareto optimal if it is not dominated
by any other solution. Therefore, the solutions set of the multi-objective
minimization problem is given by $\mathcal{P}:=\left\{ x\mid\nexists x^{\prime}:x^{\prime}\prec x\right\} $.
This set of all Pareto optimal solutions is called the Pareto set
while the collection of their images in the objective space is called
the Pareto front.
\end{definition}

\textbf{Gradient-based MOO.} \citet{desideri2012multiple}  shows that the descent direction can be found in the convex hull composed of gradients corresponding to different objectives $\{\boldsymbol{h}^{i} := \nabla f_{i}\}_{i=1}^m$:
$$\mathcal{CH} = \left\{\boldsymbol{H}^\top \boldsymbol{\alpha} = \sum_{i=1}^m \alpha_i \boldsymbol{h}^{i} \mid \alpha \in \Delta_{m} \right\}$$
 where $\Delta_m$ denotes the $m-1$ dimension simplex. 
 
 Moreover, they introduce the Multiple Gradient Descent Algorithm (MGDA), which calculates the minimum-norm gradient vector $\boldsymbol{h}$  that lies in the convex hull:   $\boldsymbol{h} = \text{argmin}_{h\in\mathcal{CH}} ||h||_2$. This approach can guarantee that the obtained solutions lie on the Pareto front, from which we cannot find an updating direction that decreases all objectives simultaneously.

% \vspace*{-1mm}
\section{Proposed method}
\label{sec:method}
% \vspace*{-1mm}

In GroupDRO, the updating formula solely focuses on the worst-performing group, which is susceptible to challenges when groups exhibit varying levels of loss magnitude, noise, and transfer characteristics. For example, among the high-loss group, there might exist a few outliers that have large input shifts or even wrong labels, which can raise the training loss of the whole group and prevent the model from fitting other groups. 
% \vspace*{-1mm}
\subsection{Balancing group distributional robustness}
\label{sec:balancing}
% \vspace*{-1mm}

\begin{figure*}[!ht]
    \centering
     \includegraphics[width=.89\textwidth, ]{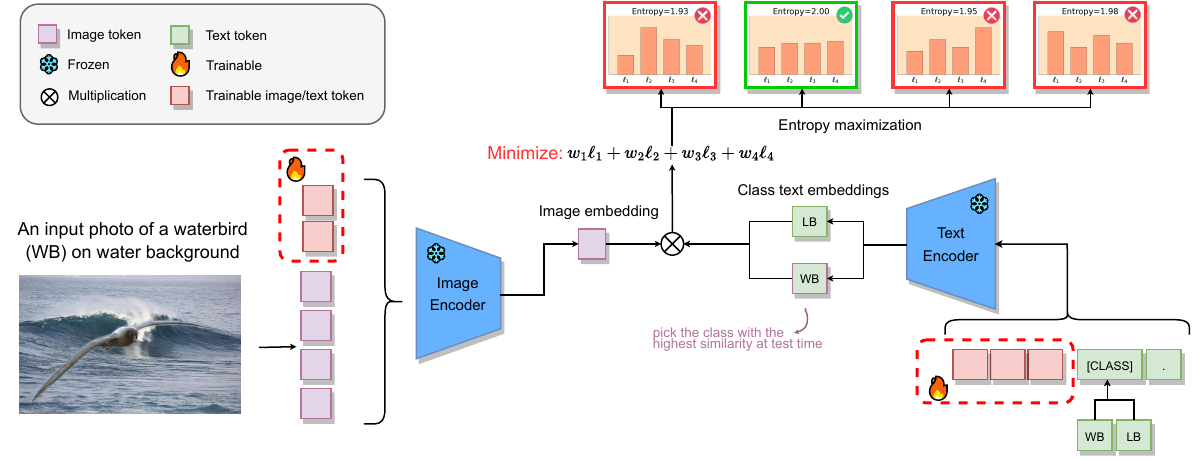}
\vspace*{-2mm}
     
    \caption{Overview of our method on the Waterbirds dataset. Our main objective is to not only improve model performance across groups by optimizing their loss functions $\ell_1, \ell_2, \ell_3, \ell_4$, but also maximize the entropy over this loss distribution.}
    \label{fig:overview}
\vspace*{-4mm}
    
\end{figure*}

In this paper, we go beyond the worst group approach of GroupDRO to examine all groups at a time. To encourage the model to learn from different groups at each step to avoid struggling on just the hardest one only, we propose to minimize the following $K$-dimension loss vector:
\begin{equation}
 \vec{\mathcal{L}}(\theta) = \big\{\mathbb{E}_{(x, y) \sim P_{g^k} (\mathcal{X}, \mathcal{Y}) }\left[\ell\left(f_\theta(x), y\right)\right] \big\}_{k=1}^K \label{eq:group_loss}
 % \vspace*{-2mm}
\end{equation}
where $K$ is the number of groups and $g^k$ is the $k$-th group.

% The above formula (\ref{eq:group_loss}), in fact, does not approximate the true loss over groups well, as it does not consider their population size. Following \cite{sagawa2019distributionally, piratla2021focus}, we modify each term to take the group size into account:
% $$\big\{\mathbb{E}_{P_{g^k} }\left[\ell\left(f_\theta(x), y\right)\right] + \frac{C}{{\sqrt{n_{g^k}}}} \big\}_{k=1}^K$$
% where $C \in \mathbb{R}^+$ is a hyperparameter, controlling the effect of group size in optimization. For brevity of notations, we omit the population size adjustment term throughout this section. 

At $t$-th iteration, we search for an update vector that minimizes the weighted sum of group losses:
% \vspace*{-2mm}
\begin{align}
d_t (w) =  w^\top  \nabla \vec{\mathcal{L}}(\theta_t) = \sum_{k=1}^K w_k \nabla \ell_k(\theta_t)
\label{eq:simplex_group_loss} 
\end{align} 
% \vspace*{-2mm} 
\\where $\ell_k(\theta)=\mathbb{E}_{P_{g^k} }\left[\ell\left(f_\theta(x), y\right)\right] $ and $w\in \Delta_{_K}$. 

Then, the updating formula of our algorithm is then given by $\theta_{t+1}=\theta_t-\eta d_t (w)$ for some step size $\eta \in \mathbb{R}^+$. The coefficient $w$ is chosen such that our update rule can benefit all groups $d_t (w)^\top \nabla \ell_k(\theta_t) \geq 0$. To this end, a straightforward approach that is readily applicable to optimize the objective (\ref{eq:group_loss}) is applying MGDA to find the solution of (\ref{eq:simplex_group_loss}) that yields the minimum Euclidean norm of the composite gradient:
\begin{align}
d_t^* =  \argmin_{w\in \Delta_{_K}} \left\|\sum_{k=1}^K w_k \nabla \ell_k(\theta_t)\right\|_2
\label{eq:mgda_solution}
\end{align}
However, MGDA is often biased toward objectives with smaller gradient magnitudes \cite{liu2021towards}, and lacks controllable property (i.e. could not adjust the group that we want to focus more or less at each update), which can be utilized to balance the loss between groups.  Hence, we introduce a new objective that steers the updating direction to obtain our desired balanced magnitude among groups. Assuming that the entropy $H(p)$ measures the diversity of a continuous distribution $p$, this behavior could be achieved via  maximization of the entropy with respect to the distribution of loss functions:
\begin{align}
\mathcal{L}_{ent}(\theta) &= H\big(\text{softmax}\big(\vec{\mathcal{L}}(\theta))\big) \nonumber\\
&= H\Big( \text{softmax}\big(\big\{\mathbb{E}_{P_{g^k} }\left[\ell\left(f_\theta(x), y\right)\right] \big\}_{k=1}^K\big)\Big)
\label{eq:entropy}    
\end{align}

\begin{theorem}
\label{thm:mimize_entropy}
Assume that the loss function $\ell$ is differentiable up to the first order with respect to $\theta$, then following 
% \vspace*{-4mm}
$$
d_{ent}:=\sum_{i=1}^{K} \nabla \ell_i(\theta) \Big[p_i\log (p_i)-p_i \sum_{j=1}^{K} \log (p_j) p_j]  \Big] 
% \vspace*{-3mm}
$$

where $p_i = \frac{e^{\ell_i (\theta)}}{\sum_{j=1}^{K}e^{\ell_j (\theta)}}$, maximizes the objective $\mathcal{L}_{ent}(\theta)$.
\end{theorem}

The proof of Theorem \ref{thm:mimize_entropy} is deterred to Appendix \ref{sec:app_proof}. Theorem \ref{thm:mimize_entropy} provides us with the updating rule that increases the entropy term in Equation (\ref{eq:entropy}) the most. To balance the learning across groups, we seek the coefficient $w$ that maximizes the cosine similarity with $d_{ent}$  and does not conflict against the gradient vector of each group. Denotes $w'_i = p_i\log (p_i)-p_i \sum_{j=1}^{K} \log (p_j) p_j$, this is equivalent to optimizing:
% \vspace*{-2mm}
\begin{align}
\label{eq:entropy_similarity}
w^*  &= \argmax_{w \in \Delta_K} d_t (w)^\top d_{ent} \\
&  \text{s.t.}\;\;\;  d_t (w)^\top \nabla \ell_k(\theta_t) \geq 0 \; \forall \, k \in [K]. \nonumber
\end{align}

% Motivated by prior work on constrained multi-objective optimization \cite{lin2019pareto, momma2022multi, liu2021conflict}, we define the index set of all activated constraints as:
% $$\mathcal{S}\left(\theta_t\right):=\left\{k \mid d_{ent}^\top \nabla l_k(\theta_t) < 0 , k \in [K] \right\}.$$
% i.e. $w'^\top \nabla \vec{\mathcal{L}}(\theta)^\top \nabla l_k(\theta_t) < 0 \;\forall\; k \in \mathcal{S}\left(\theta_t\right)$.
% \vspace*{-2mm}

While the balance between loss magnitude is not achieved, we strictly follow the updating direction that decreases training errors of high-loss groups, but slightly relax constraints in the optimization problem (\ref{eq:entropy_similarity}) to allow bounded increases among groups that already have small loss magnitudes.
% \vspace*{-2mm}
{\begin{align}
\label{eq:main_objective}
&w^*  = \argmax_{w \in \Delta_K} d_t (w)^\top d_{ent} \nonumber \\
&\;\;\;\;= \argmax_{w \in \Delta_K} w^\top  \nabla \vec{\mathcal{L}}(\theta)^\top  \nabla \vec{\mathcal{L}}(\theta)  w' \\
&\text{s.t.}\;\;\;  \forall \, k \in [K]:\nonumber\\
& \;\;d_t (w)^\top \nabla \ell_k(\theta_t) \geq 0 \; \text{if}\;\ell_k(\theta_t) = \max_{k\in[K]} \ell_k(\theta_t). \nonumber\\
 & \;\; d_t (w)^\top \nabla \ell_k(\theta_t) \geq d_{ent}^\top \nabla \ell_k(\theta_t)  \;\text{if}\;\ell_k(\theta_t) < \max_{k\in[K]} \ell_k(\theta_t). \nonumber
 % \vspace*{-2mm}
\end{align}
 }{
The second constraint guarantees that the obtained solution is not worse than $d_{ent}$ on those groups in which the model is performing well or when $d_{ent}$ itself intrinsically supports learning them. Once the loss magnitude is almost balanced ($||d_{ent}||\leq \epsilon $), we simply update the model parameter by the average group gradients, i.e. $w=(\frac{1}{K},\frac{1}{K}, \dots, \frac{1}{K})$. }

In summary, the key difference between our proposed method and GroupDRO is that we solve a small $K$-dimensional optimization problem to find the updating direction that not only focuses on the worst group but also improves the performance in other groups and balances the loss magnitude between them at the same time. In practical settings, the group number $K$ is often a small number (e.g. $K=4$), thus could be efficiently solved using standard linear programming libraries \cite{diamond2016cvxpy, andersen2020cvxopt} without introducing significant computational overhead.

% \vspace*{-1mm}
\subsection{Controllable prompt tuning for balancing group distributional robustness}
% \vspace*{-1mm}

To achieve the controllable property, we apply the optimization procedure in section \ref{sec:balancing} on $\ell_k(\theta)={c}_k\mathbb{E}_{P_{g^k} }\left[\ell\left(f_\theta(x), y\right)\right] $ where $\mathbf{c}\in\mathbb{R}^K$ is a pre-defined vector that is used to adjust the magnitude of group losses. Intuitively, the principle behind our proposed objective is to enforce the loss vector $\vec{\mathcal{L}}(\theta)$ to be inversely proportional to $\mathbf{c}$, from which we can adjust the group loss magnitude, and thus control the trade-off between them by varying $\mathbf{c}$. For example, we can put more weight on penalizing loss terms of some specific groups to accelerate the learning progress on those groups, or conversely slow down the learning to avoid overfitting on oversampled groups \cite{deng2023robust}.

In practice, tuning the value of $\mathbf{c}$ to obtain our desired model behavior requires learning multiple models individually and independently. Therefore, the computational complexity of our proposed method can grow linearly with the number of vectors $\mathbf{c}$, thus being very expensive in the context of deep learning. To reduce overhead, we adopt the usual practice of parameter-efficient fine-tuning methods and optimize only a small portion of the model. In particular, we freeze the main backbone while introducing lightweight prompt sets, one for each task.  For example, we employ the same ViT backbone for the two datasets Waterbirds, CelebA,  but using different sets of prompts (Section \ref{sec:experiment}).

\textbf{Overview of our proposed method} is depicted in Figure \ref{fig:overview}, given an input image belonging to one of a group $\{g^1, g^2, \dots, g^K\}$ and a set of classes, we solve the optimization problem (\ref{eq:main_objective}) to find the reweighting coefficient $w$ that maximizes the entropy $\mathcal{L}_{ent}(\theta)$ while reasonably decreases group losses. Please note that while Figure \ref{fig:overview} illustrates our proposed method on the CLIP model on the image classification task, it is applicable to other transformer-based models and task types. In our experiment, we exploit both image-end prompts and language-end prompts tuning on frozen ViT and CLIP backbones, respectively.

\textbf{Prompt tuning for Vision Transformer}: Regarding the prompt design for a ViT of $N$ layers, we use the set of continuous learnable tokens $\{\mathbf{P}_n\}_{n=0}^{N-1}$. Each $\mathbf{P}_n \in \mathbb{R}^{L \times D}$, where  $L$, $D \in \mathbb{N}$ indicate the prompt length and latent space dimension, respectively. Subsequently, the input at the $(n+1)$-th layer follows the format of $\left[x_{n}, \mathbf{P}_n, \mathbf{E}_n\right]$, where $\mathbf{x}_n, \mathbf{E}_n \in \mathbb{R}^D$ represent the {[CLASS]} token and image patch embeddings after $n$-th layer while $[\cdot, \cdot]$  denotes the concatenation operator. Thus, the number of trainable parameters for this image-end prompt set is $N \times L \times D$.

\textbf{Prompt tuning for CLIP}: Being trained on an enormous amount of image-text pairs, CLIP offers an expressive representation for both language and image inputs by using two separate encoders, one for each modality, as shown in Figure \ref{fig:overview}. Since the image encoder can be either ResNet \citep{he2016deep}, ConvNeXt \citep{liu2022convnet} or ViT \cite{dosovitskiy2020image}, we conduct prompt tuning on the transformer text encoder for consistency. In short, we employ a learnable context prompt of length $L$:  $\mathbf{P} \in \mathbb{R}^{L \times D}$, to replace the hand-crafted prompt used by the original CLIP (i.e., ``this is
a photo of a "). Based on this design, the input for the text encoder of each class has the format of $\left[x, \mathbf{P} \right]$, where $x$ is the {[CLASS]} embedding of the corresponding class name and has the same dimension of $D$. These learnable tokens $\mathbf{P}$ are better at capturing domains-specific knowledge than those artificial prompt \citep{zhou2022learning, zhu2023prompt} and can be learned to de-bias the original CLIP model while introducing a few trainable parameters ($L \times D$).

% \vspace*{-1mm}
\section{Experiments}
\label{sec:experiment}
% \vspace*{-1mm}
In this section, we evaluate the effectiveness of the proposed CPT method on benchmark image datasets in the presence of spurious features: Waterbirds \cite{sagawa2019distributionally}, CelebA \cite{liu2015faceattributes}, MetaShift \citep{liang2021metashift} and ISIC \citep{codella2019skin}.  Due to the space limit constraint, we briefly provide an overview of our experimental setup below, see Appendix \ref{sec:app_exp} for more details and additional results.  Our implementation is available at \url{https://github.com/VietHoang1512/CPT}.

\begin{table*}[t]
% \vspace*{-1mm}
\setlength{\tabcolsep}{12pt}
  \caption{Overall results on Waterbirds and CelebA datasets with best methods are highlighted in \textbf{bold}. Performance is evaluated on the test set with models early stopped at the highest worst-group accuracy on the validation set. %We use $\surd$ or $\surd \surd$ to denote methods using group information for validation or both training and validation, respectively.
% \vspace*{-2mm}
  }
  \label{tab:wb_celeba}
  \centering
  \resizebox{.98\textwidth}{!}{%
  \begin{tabular}{lcccccccccc}
    \toprule
    & \multirow{2}{5.em}{\centering{\textbf{Train \\ group info}}} & \multirow{2}{5.em}{\centering{\textbf{Validation \\ group info}}}  & \multirow{2}{3.em}{\centering{{Train}\\{once}}} &\multicolumn{3}{c}{Waterbirds}&  \multicolumn{3}{c}{CelebA}\\
    \cmidrule(r){5-10}
     Method & & &  & Worst & Average & \# Params &  Worst & Average  & \# Params \\
    \midrule
    ERM & $\times$ & $\times$ & $\surd$ & $70.0_{\pm 2.3}$ & $97.1_{\pm 0.1}$ & $23$M & $45.0_{\pm 1.5}$ & $94.8_{\pm 0.2}$  &  $23$M  \\   
    AFR & $\times$ & $\times$ & $\times$ & $90.4_{\pm 1.1}$ & $94.2_{\pm 1.2}$ & $23$M & $82.0_{\pm 0.5}$ & $91.3_{\pm 0.3}$  &  $23$M  \\
\midrule
    LfF & $\times$ & $\surd$ & $\times$ & $78.0_{N/A}$ & $91.2_{N/A}$ & $23$M & $77.2_{N/A}$ & $85.1_{N/A}$  &  $23$M \\
    % EIIL & $\surd$ & $\times$ & $77.2_{\pm 1.0}$ & $96.5_{\pm 0.2}$ & $23$M & $81.7_{\pm 0.8}$ & $85.7_{\pm 0.1}$ &  $23$M \\
    SSA & $\times$ & $\surd$ & $\times$ & $89.0_{\pm 0.6}$ & $92.2_{\pm 0.9}$ & $23$M & $89.8_{\pm 1.3}$ & $92.8_{\pm 1.3}$  &  $23$M \\
    JTT & $\times$ & $\surd$ & $\times$ & $86.7_{N/A}$ & $93.3_{N/A}$ & $23$M & $81.1_{N/A}$ & $88.0_{N/A}$  &  $23$M \\
    PGI & $\times$ & $\surd$ & $\times$ & ${{79.5}}_{\pm 1.9}$ & $95.5_{\pm 0.8}$ & $23$M & $85.3_{\pm 0.3}$ & $87.3_{\pm 0.1}$  &  $23$M \\
    CIM & $\times$ & $\surd$ & $\times$ & ${{77.2}}_{N/A}$ & $95.6_{N/A}$ & $23$M & $83.6_{N/A}$ & $90.6_{N/A}$  &  $23$M \\
    CnC & $\times$ & $\surd$ & $\times$ & $88.5_{\pm 0.3}$ & $90.9_{\pm 0.1}$ & $23$M & $88.9_{\pm 1.3}$ & $88.9_{\pm 0.5}$  &  $23$M \\
    DFR & $\times$ & $\surd$ & $\times$ & $92.9_{\pm 0.2}$ & $94.2_{\pm 0.4}$ &  $23$M & $88.3_{\pm 1.1}$ & $91.3_{\pm 0.3}$ &  $23$M \\   
    \midrule
    UW & $\surd $ & $ \surd$ & $\surd$ & $88.0_{\pm 1.3}$ & $95.1_{\pm 0.3}$ & $23$M & $83.3_{\pm 2.8}$ & $92.9_{\pm 0.2}$ &   $23$M \\ 
    Subsample & $\surd $ & $\surd$ & $\surd$ & $86.9_{\pm 2.3}$ & $89.2_{\pm 1.2}$ & $23$M & $86.1_{\pm 1.9}$ & $91.3_{\pm 0.2}$ &   $23$M \\ 
    
    LISA & $\surd$ & $ \surd$ & $\times$ & $89.2_{\pm 0.6}$ & $91.8_{\pm 0.3}$ &  $23$M & $89.3_{\pm 1.1}$ & $92.4_{\pm 0.4}$ &  $23$M \\
    GroupDRO & $\surd$ &  $\surd$ & $\surd$ & $86.7_{\pm 0.6}$ & $93.2_{\pm 0.5}$ & $23$M & $86.3_{\pm 1.1}$ & ${92.9}_{\pm 0.3}$ &  $23$M \\
    CGD & $\surd$ & $ \surd$ & $\surd$ & $88.9_{\pm 0.8}$ & $91.3_{\pm 0.6}$ & $23$M & $90.0_{\pm 0.8}$ & ${92.5}_{\pm 0.2}$ &  $23$M \\
        DISC & $\surd$ &  $\surd$ & $\times$ & ${{88.7}}_{\pm 0.4}$ & $93.8_{\pm 0.7}$ & $23$M & N/A & N/A & N/A \\
    PDE & $\surd $ &  $\surd$ & $\surd$ & ${{90.3}}_{\pm 0.3}$ & $92.4_{\pm 0.8}$ & $23$M & ${{91.0}}_{\pm
0.4}$ & $92.0_{\pm 0.6}$ & $23$M \\
\midrule
    CPT & $\surd $ &  $\surd$ & $\surd$ & ${\mathbf{93.5}}_{\pm 0.4}$  & ${{96.3}}_{\pm
0.2}$ & $\mathbf{94}$k & $\mathbf{92.0}_{\pm 0.3 }$ & ${93.2}_{\pm 0.3}$ & $\mathbf{94}$k  \\
    \bottomrule
  \end{tabular}
  }
\vspace*{-3mm}
\end{table*}
% \vspace*{-1mm}
\subsection{Datasets}
% \vspace*{-1mm}
\textbf{Waterbirds}  is created by placing bird photos from the Caltech-UCSD Birds dataset \citep{wah2011caltech} with background images taken from Places \citep{zhou2017places}. The target attributes are foreground objects (i.e. landbird or waterbird) while spurious correlations are background (i.e. land or water landscape).  \textbf{CelebA} is a factual dataset containing $200$K face images of celebrities. The goal is to classify blond/non-blond hair color. Statistically, more than 94$\%$ of blond hair examples in the training set are women.

\textbf{MetaShift} is another real-world dataset for cat-dog classification, where the cat and dog images are spuriously correlated with indoor (e.g. bed) and outdoor objects (e.g. bike). For evaluation, models are given dog and cat images associated with shelf backgrounds, which do not appear in the training set. Thus, this dataset represents a more coherent distributional shift, compared to other benchmarks. \textbf{ISIC} is an even more challenging dataset with the appearance of multiple spurious features  \citep{codella2019skin}, which includes dermoscopic images of skin lesions with target attributes benign or melanoma. Visual examples of those datasets, along with the size for each group in train, validation, and test sets are given in Appendix \ref{sec:app_exp}.  
% \vspace*{-1mm}
\subsection{Baselines} 

Following the previous studies, we compare CPT against other state-of-the-art methods for mitigating spurious correlations, including DFR \cite{kirichenko2022last}, GroupDRO \cite{sagawa2019distributionally}, PDE \citep{deng2023robust}, Subsample \citep{deng2023robust}, CGD \citep{piratla2021focus} with the access to the group information during training. Besides, we consider those methods require group labels for validation:  LfF \cite{nam2020learning}, JTT \citep{liu2021just}, PGI \citep{ahmed2020systematic}, CIM \citep{taghanaki2021robust}, CNC \citep{zhang2022correct}, or those do not use group annotations at all like ERM and AFR  \citep{qiu2023simple}. Moreover, we also report the results from two recent studies of spurious correlation on ViT \citep{ghosal2023vision} and CLIP \citep{ yang2023mitigating, dehdashtian2024fairerclip}. 

Regarding the baselines for those experiments further taking the distribution shift into account, we follow exactly the protocol of \citet{wu2023discover} and compare our method to Upweighting (UW), IRM \citep{arjovsky2019invariant}, IB-IRM \citep{ahuja2021invariance}, V-REx \citep{krueger2021out}, CORAL \citep{sun2017correlation}, and Fish \citep{shi2021gradient};
instance reweighting methods: JTT \citep{liu2021just},
DM-ADA \citep{xu2020adversarial},  LISA \cite{yao2022improving} and the current state-of-the-art method, DISC \citep{wu2023discover}.
% \vspace*{-1mm}
\subsection{Model training and evaluation}
% \vspace*{-1mm}
We examine our proposed method on ResNet50 \citep{he2016deep}, ViT-B/16 \cite{dosovitskiy2020image} and different variants of OpenAI's CLIP \citep{radford2021learning} to align with previous work \citep{kirichenko2022last, ghosal2023vision, yang2023mitigating}. For prompt tuning on ViT, we employ prompts of lengths $L=5$ in the MetaShift and ISIC, or $L=10$ in the Waterbirds and CelebA datasets. Similarly, $L$ is set to $16$ for all CLIP backbones throughout our experiment. Unless otherwise explicitly mentioned in the experiment, the value of the controlling vector $c$ is set to the default value of $(1,\dots, 1)$. Other detailed configurations are relegated to Appendix \ref{sec:app_exp}.

Due to the large number of groups in ISIC, we calculate the AUROC score \cite{bradley1997use}, following \citep{wu2023discover, bissoto2020debiasing} while computing average, worst-group and mean (if possible) accuracy scores for other datasets. For a fair and precise comparison, results of baselines are taken from recent studies \citep{deng2023robust, wu2023discover, piratla2021focus, zhang2022correct, kim2023removing} and reported directly from original papers.
% \vspace*{-1mm}
\subsection{Results}
% \vspace*{-1mm}
\textbf{ Waterbirds and CelebA.} We start by validating CPT on two well-established benchmarks for investigating spurious correlation, namely Waterbirds and CelebA in Table \ref{tab:wb_celeba}.  From the results, we see that CPT consistently achieves the best worst-group accuracy compared with the recent state-of-the-art methods such as DFR and PDE on both datasets. Remarkably, our method only updates a tiny portion ($0.4\%$) of parameters. As such, CPT is much more computationally efficient than methods that need to update the entire network. Furthermore, we see that CPT has strong gains in average accuracy, thus not only outperforming other spurious correlation methods on both metrics but also closing the gap to the average accuracy of ERM. Another interesting observation is that those baselines aim to balance the learning between groups are effective in alleviating spurious correlation, e.g. via downsampling the majority groups (Subsample), reweighting group losses inversely proportional to the group populations (UW), or computing scaling weights using group training performance to upweight the worst-group loss (GroupDRO). Even though, CPT still exhibits impressive performance compared to the aforementioned methods by using an adaptive balancing mechanism at each iteration instead of relying on fixed reweighting coefficients.

% \vspace*{-3mm}
\begin{figure}[!ht] 
    \centering
     \includegraphics[width=1\columnwidth]{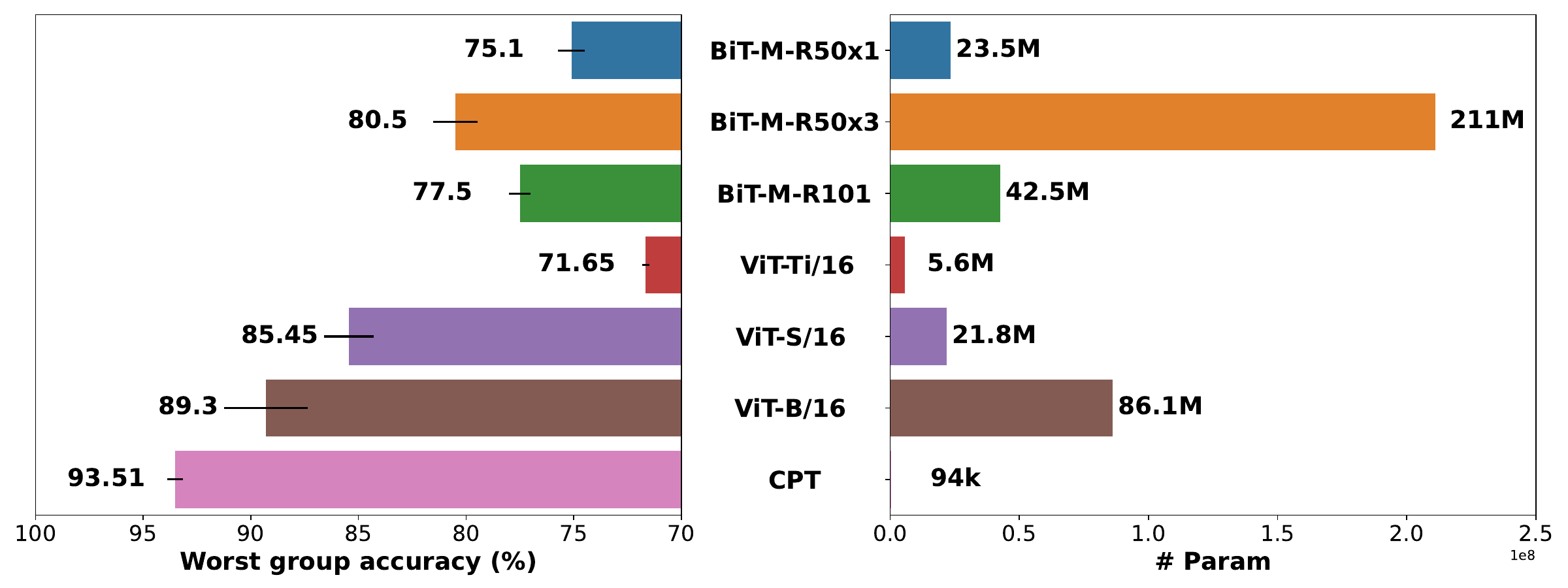}
     % \vspace*{-2mm}
    \caption{Results of fine-tuning ViT backbones on
Waterbirds. Error bars represent the standard deviation over independent runs.}
    \label{fig:vit-variant}
\end{figure}
% \vspace*{-4mm}

\textbf{De-bias ViT and CLIP.} Motivated by recent studies focusing on investigating spurious correlations of pre-trained transformer-based models \citep{yang2023mitigating, dehdashtian2024fairerclip}, we apply our proposed algorithm on ViT and CLIP and then compare them against those work. It is worth noting that, the authors did not experiment on CelebA with the blond hair target. Hence, we only show their original results on Waterbirds in Figure \ref{fig:vit-variant} and Table \ref{tab:clip_wb}. As can be seen in Figure \ref{fig:vit-variant}, while CPT enjoys much less computation and memory overhead (only $0.1\%$ and $0.04\%$ compared to ViT-B/16 itself and BiT-M-R50x3 \citep{ghosal2023vision}) it still outperforms the second-best model by a large margin. This significant improvement comes from our prompting design and optimization contribution, respectively. Details of backbones are given in Appendix \ref{sec:app_exp}.

\begin{table}[t]
\caption{Performance of CLIP models on Waterbirds. $\dagger$ indicates our implemented results. We use \textbf{Bold} font to indicate the best methods for different numbers of trainable parameters. N/A denotes those methods do not report the number of trainable parameters. \label{tab:clip_wb}}
% \vspace*{-2mm}
  \centering
  \resizebox{1\columnwidth}{!}{\begin{tabular}{l|ccc|ccc}
\toprule Model & \multicolumn{3}{c|}{ ResNet50 } & \multicolumn{3}{c}{ ViT-L/14@336px } \\
& Average & Worst & \# Params & Average & Worst & \# Params \\
\midrule Pre-trained CLIP & ${9 0 . 8 }$ & $44.9 $ & $0$ & $88.5 $ & $34.0 $ & $0$ \\
Fine-tuned CLIP & $81.3 $ & ${7 7 . 1 }$ & $14789632$ & ${9 7 . 2 }$ & $89.7 $ & $786432$ \\
\midrule ERM & ${93 . 5 }$ & $54.4 $ & $14789632$ & $96.8 $ & $58.1 $ & $786432$ \\

ERM Adapter & $96.0 $ & ${63.0 }$  & $524288$ & $97.8 $ & $76.1 $ & $524288$ \\

Contrastive Adapter & $88.2 $ & ${82.5 }$  & $263424$ & $94.5 $ & $85.3$ & $197632$ \\
DFR & $91.8 $ & $63.9$  & N/A & $96.1 $ & $65.9$ & N/A \\
FairerCLIP & $84.3 $ & $75.4$  & N/A & $92.2$ & $86.0$  & N/A \\
\midrule
GroupDRO & $83.3 $ & $73.7 $ & $14789632$ & $94.1 $ & ${9 0 . 8 }$  & $786432$ \\
\citet{yang2023mitigating} & $83.2 $ & ${7 7 . 5 }$ & $14789632$ & $96.9 $ & ${9 0 . 5 }$  & $786432$ \\
CPT$_{\text{balance}}$  & $85.8$ & $\mathbf{82.8}$ & $14789632$ & $94.0$ & $\mathbf{92.9}$  & $786432$ \\

\midrule
ERM$^\dagger$ & ${85.7}$ & $64.6$ & $8192$ & $97.4$ & $84.3$  & $12288$ \\

GroupDRO$^\dagger$ & $81.0$ & $76.7$ & $8192$ & $95.8$ & $90.5$  & $12288$ \\

CPT & $83.3$ & $78.4$ & $8192$ & $97.0$ & $90.2$  & $12288$ \\
CPT$_{\text{balance}}$ & $81.8$ & $\mathbf{79.8}$ & $8192$ & $95.6$ & $\mathbf{91.9}$  & $12288$ \\
\toprule
\end{tabular}}
\vspace*{-5mm}
\end{table}

As shown in Table \ref{tab:clip_wb}, CPT$_{\text{balance}}$ (i.e. CPT with $c=(1,\dots, 1)$) clearly outperforms all other alternatives using the same number of trainable parameters, which proves the effectiveness of our optimization framework. Furthermore, when varying the coefficient $c$, one can control the trade-off between group performance, yielding a much more flexible learning algorithm for group distributional robustness. Indeed, increasing coefficients of the majority groups (e.g. $c=(2, 1, 1, 1.5)$ in this setting) helps improve the average accuracy of CPT to perform on par or better than \citet{yang2023mitigating} on both metrics while updating only $0.05\%$ and $1\%$ of the model parameters. Thus, this again suggests the clear efficiency advantage of our prompt tuning, which offers scalability for controlling adjustment. Since CPT has a larger capacity to control the learning between groups,  we can sacrifice worst-group accuracy for the gain in performance on other groups and vice-versa (last two rows). Intriguingly, while obtaining impressive performance on CLIP-RN50, Contrastive Adapter \citep{zhang2022contrastive} falls far behind other baselines on CLIP ViT-L/14@336px.

\vspace*{-2mm}
\begin{table}[!ht]
  \caption{Experimental results on MetaShift and ISIC dataset.}
  % \vspace*{-2mm}
  \label{tab:meta_isic}
  \centering
  \resizebox{.9\columnwidth}{!}{%
\begin{tabular}{l|cc|c|c}
\toprule
     &\multicolumn{2}{c|}{MetaShift}&  {ISIC}\\
    \cmidrule(r){2-4}
     Method   & Average  & Worst & AUROC  & \# Params \\
\toprule ERM & $72.9_{\pm 1.4} $ & $62.1_{\pm 4.8} $  & $36.4_{\pm 0.7} $  &   $23$M \\
UW & $72.1_{\pm 0.9} $ & $60.5_{\pm 3.8 }$ & $39.2_{\pm 0.6} $  &   $23$M \\
IRM & $73.9_{\pm 0.8} $ & $64.7_{\pm 2.1 }$ & $45.5_{\pm 3.6} $ &   $23$M \\
IB-IRM & $74.8_{\pm 0.2} $ & $65.6_{\pm 1.1} $  &  $38.6_{\pm 1.7} $ &   $23$M \\
V-REx & $72.7_{\pm 1.7} $ & $60.8_{\pm 5.5} $ &  $24.5_{\pm 6.4} $  &   $23$M  \\
CORAL & $73.6_{\pm 0.4} $ & $62.8_{\pm 2.7} $ &  $37.9_{\pm 0.7} $  &   $23$M\\
Fish & $64.4_{\pm 2.0} $ & $53.2_{\pm 4.5} $ &  $42.0_{\pm 0.8} $ &   $23$M\\
GroupDRO & $73.6_{\pm 2.1} $ & $66.0_{\pm 3.8 }$ &  $36.4_{\pm 0.9} $ &   $23$M\\
JTT & $74.4_{\pm 0.6} $ & $64.6_{\pm 2.3} $ & $33.8_{\pm 0.0} $  &   $23$M\\
DM-ADA & $74.0_{\pm 0.8} $ & $65.7_{\pm 1.4 }$ &   $35.8_{\pm 1.0} $  &   $23$M\\
LISA & $70.0_{\pm 0.7} $ & $59.8_{\pm 2.3} $ &  $38.0_{\pm 1.3} $ &   $23$M\\
DISC & $75.5_{\pm 1.1} $ & $73.5_{\pm 1.4} $ &  $55.1_{\pm 2.3} $ &   $23$M\\
\midrule CPT & $ \mathbf{79.8}_{\pm 2.6} $ & $\mathbf{77.0}_{\pm{0.8} }$  & $ \mathbf{64.9}_{\pm 1.2} $  &   $\mathbf{47}$k\\
\bottomrule
\end{tabular}}
\end{table}
\vspace*{-1mm}

\textbf{MetaShift and ISIC.} Results on those datasets are depicted in Table \ref{tab:meta_isic}, along with the number of parameters required for each baseline. We mostly pick those that utilize group information during training for better benchmarking the efficiency of our proposed framework. In summary, the performance of CPT exceeds other methods by large margins, especially $\approx 10$ AUROC score on ISIC. On MetaShift, it is also the best method in terms of average and worst-group accuracy, improving the performance by approximately $4\%$.

\subsection{Ablation studies}
In this section, we provide different ablation studies to see how each component in our proposed method individually contributes to the overall improvement of CPT.

\textbf{Optimization and Backbones.} In Table \ref{tab:ablation}, CPT$_{\text{w/o entropy}}$ eliminates the entropy objective and uses the average gradient $1/K \sum \nabla \ell_i(\theta)$ to update the model. While this strategy can improve mean accuracy among groups, compared to ERM, the performance has a significant drop compared to our optimization procedure. This conclusion suggests that conducting simple gradient averaging when there are still large differences among group losses could cause severe bias in training. On the other hand, there is no significant difference between the performance of fully fine-tuning ResNet50  and prompt tuning on ViT-B/16, which indicates that our method does not benefit much from more powerful backbones. Instead, we remind readers that prompting in CPT is introduced to scale up our proposed algorithm. Thus, even applying our proposed balancing method alone can boost the performance of ResNet to exceed GroupDRO by large margins on WaterBirds $(5.2\%)$ and CelebA $(5.1\%)$.

\begin{table}[!ht]
% \vspace*{-2mm}
  \caption{Ablation results for different architectures and optimization procedures.}
  % \vspace*{-2mm}
  \label{tab:ablation}
  \centering
  \resizebox{0.95\columnwidth}{!}{%
\begin{tabular}{llccc}
\toprule
Dataset & Method & Average & Worst   & \# Params   \\
\toprule  \multirow{6}{*}{MetaShift} & ERM & $72.9_{\pm 1.4} $ & $62.1_{\pm 4.8} $  &   $23$M \\
 & GroupDRO & $73.6_{\pm 2.1} $ & $66.0_{\pm 3.8 }$  &   $23$M\\
& DISC & $75.5_{\pm 1.1} $ & $73.5_{\pm 1.4} $  &   $23$M\\
& CPT$_{\text{w/o entropy}}$ (ResNet50) & $ {76.1}_{\pm 0.7} $ & ${62.8}_{\pm 1.8 }$  &    $23$M\\
& CPT  (ResNet50) & $ {78.2}_{\pm 2.5} $ & ${76.9}_{\pm 2.0 }$  &    $23$M\\
& ERM (prompt-tuning)  & $78.4_{\pm 1.4}$ & $65.0_{\pm 2.4}$ & $\mathbf{47}$ \\
& CPT$_{\text{w/o entropy}}$ & ${77.3}_{\pm0.8 }$  & $ {74.7}_{\pm 2.5 } $  &   $\mathbf{47}$k\\
& CPT & $ \mathbf{79.8}_{\pm 2.6} $ & $\mathbf{77.0}_{\pm{0.8} }$  &   $\mathbf{47}$k\\
\midrule
\multirow{4}{*}{Waterbirds} & ERM & $97.1_{\pm 0.1}$ & $70.0_{\pm 2.3}$  & $23$M \\
& GroupDRO  & $93.2_{\pm 0.5}$ & $86.7_{\pm 0.6}$  & $23$M  \\
 &   CPT (ResNet50)  & ${90.9}_{\pm
0.2}$ & ${91.9}_{\pm 0.3}$   & $23$M\\ 
& ERM (prompt-tuning)  & $97.5_{\pm 0.2}$ & $64.4_{\pm 2.6}$ & $\mathbf{94}$k \\
 &   CPT  & ${{96.3}}_{\pm
0.2}$  & ${\mathbf{93.5}}_{\pm 0.4}$  & $\mathbf{94}$k \\
\midrule
\multirow{4}{*}{CelebA} & ERM & $94.8_{\pm 0.2}$ &   $45.0_{\pm 1.5}$   &  $23$M \\
&GroupDRO  & ${92.9}_{\pm 0.3}$ & $86.3_{\pm 1.1}$ &  $23$M \\
 & CPT (ResNet50) & ${92.7}_{\pm 0.3}$ & ${91.4}_{\pm 0.8 }$  & $23$M  \\
 & ERM (prompt-tuning)  & $95.6_{\pm 0.3}$ & $58.3_{\pm 1.9}$ & $\mathbf{94}$k \\
&CPT & ${93.2}_{\pm 0.3}$ & $\mathbf{92.0}_{\pm 0.3 }$  & $\mathbf{94}$k  \\
 
\bottomrule
\end{tabular}}
\vspace*{-3mm}
\end{table}

\begin{figure}[!ht] 
    \centering
    \vspace*{-3mm}
     \includegraphics[width=1\columnwidth]{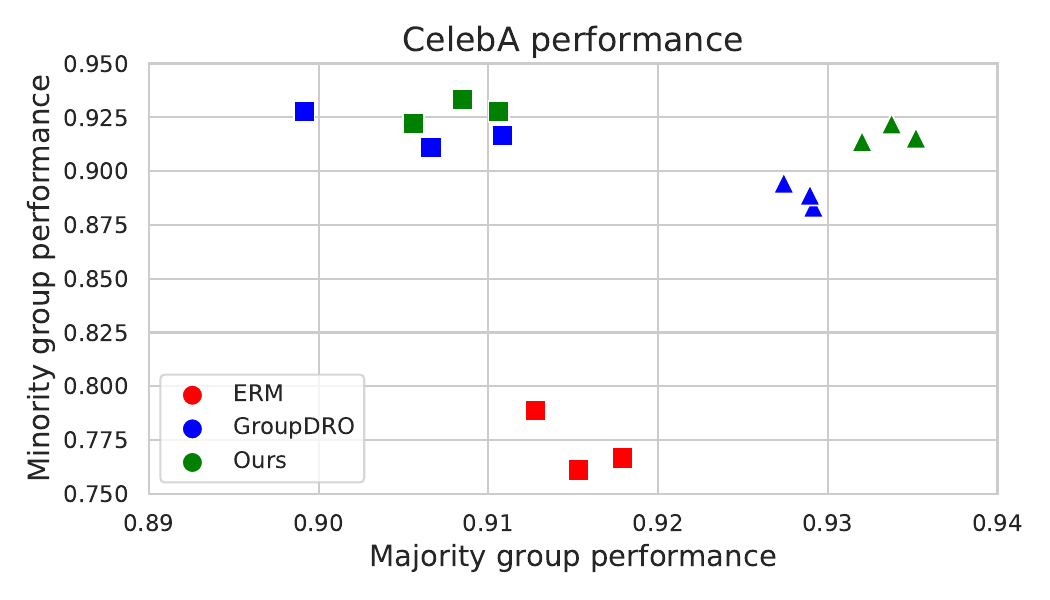}
     \vspace*{-2mm}
    \caption{Performance of ResNet50 at: the last checkpoint used for evaluation (highest worst group accuracy on validation set), denoted by $\bigtriangleup$, and the checkpoint where the performance on the minority group is highest, denoted by  $ \square$. Results are obtained on three random seeds.}
    \label{fig:celeba-gap}
    % \vspace*{-2mm}
\end{figure}

\textbf{Gap among group performance.} Apart from Figure \ref{fig:acc_curve}, we visualize the performance of each comparative method at its early stopping epoch and the point when it actually obtains the highest score on the minority group in Figure \ref{fig:celeba-gap}. Those points collapse in the case of ERM since it does not have any balancing mechanism. Unsurprisingly, the gap between minority and majority performance of ERM is largest ($>10\%$) while this figure decreases for GroupDRO and CPT. It is true that GroupDRO can obtain a relatively high score on the minority group early in training, even on par with CPT. However, its majority group performance at this point has not converged yet ($\approx90.5\%$), and when converged, its performance on the minority group declines considerably ($91\%\rightarrow88\%$). By contrast, the is only a small difference in terms of minority performance for CPT at those points, allowing to have equally good performance across groups through training.

\textbf{Effect of controlling vector} Table \ref{tab:upweight} presents the distribution of the CelebA dataset and the performance of GroupDRO and our proposed method with different values of the controlling vectors. We can see that setting  $c$ equal to the default value $(1,1,1,1)$ can perfectly balance the accuracy score across groups while GroupDRO encounters difficulty in fitting the minority one. Furthermore, by varying the value of $c$, we could observe different behavior of the model in each group. For example, the accuracy scores on two smaller groups ($2 \& 3$) increase significantly, as a result of the loss magnitudes decrease when we put more weight on penalizing those two. With a tiny sacrifice on the performance of two majority groups, the model could gain considerable improvement in minority groups and the overall performance.

% \vspace*{-2mm}
\begin{table}[!ht]
\caption{Performance of the model after training for one epoch. Here, we simply adjust the optimization in favor of two minority groups to observe their improvement. \label{tab:upweight} }
% \vspace*{-2mm}
\resizebox{1\columnwidth}{!}{% 
\centering
\begin{tabular}{c|c|c|c|c||c|c|c}
\toprule
    \textbf{Group $g$ } (size) & \textbf{0} $(44\%)$& \textbf{1} $(41\%)$& \textbf{2} $(14\%)$ & \textbf{3} $(1\%)$  & Worst &  Avg &  Mean \\
\midrule
GroupDRO & $92.0$  & $92.5$ & $91.5$ & $86.1$ & $86.1$ & $92.08$ & $89.4$    \\
\midrule
CPT $c=(1,1,1,1)$ & $\mathbf{92.9}$  & $\mathbf{93.4}$ & $92.8$ & $91.7$ & $\mathbf{91.7}$ & $\mathbf{93.1}$ & $92.7$\\
CPT $c=(1,1,1,2)$ & $92.0$  & $91.1$ & $93.1$ & $\mathbf{95.6}$ & $91.1$ & $91.9$ & $93.0$\\
CPT $c=(1,1,2,1)$ & $91.2$  & $91.8$ & $\mathbf{95.2}$ & $95.0$ & $91.2$ & $92.0$ & $\mathbf{93.3}$\\
\midrule
    \end{tabular}
}
% \vspace*{-4mm}
\end{table}

\begin{figure}[!ht]
    \centering
     \includegraphics[width=1\columnwidth, ]{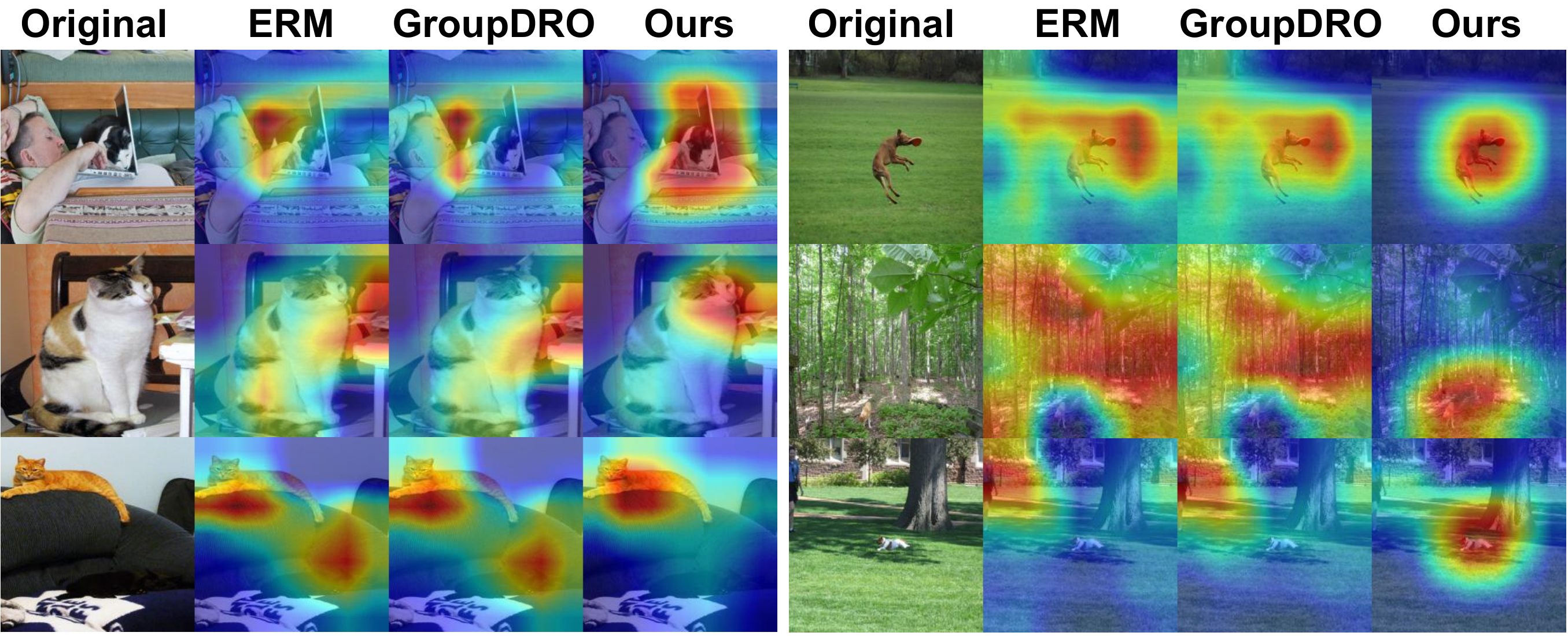}
    \caption{GradCAM of ResNet50 on in-domain samples.}
    \label{fig:gradcam-id}
    \vspace*{-3mm}
\end{figure}

\begin{figure}[!ht]
    \centering
     \includegraphics[width=1\columnwidth, ]{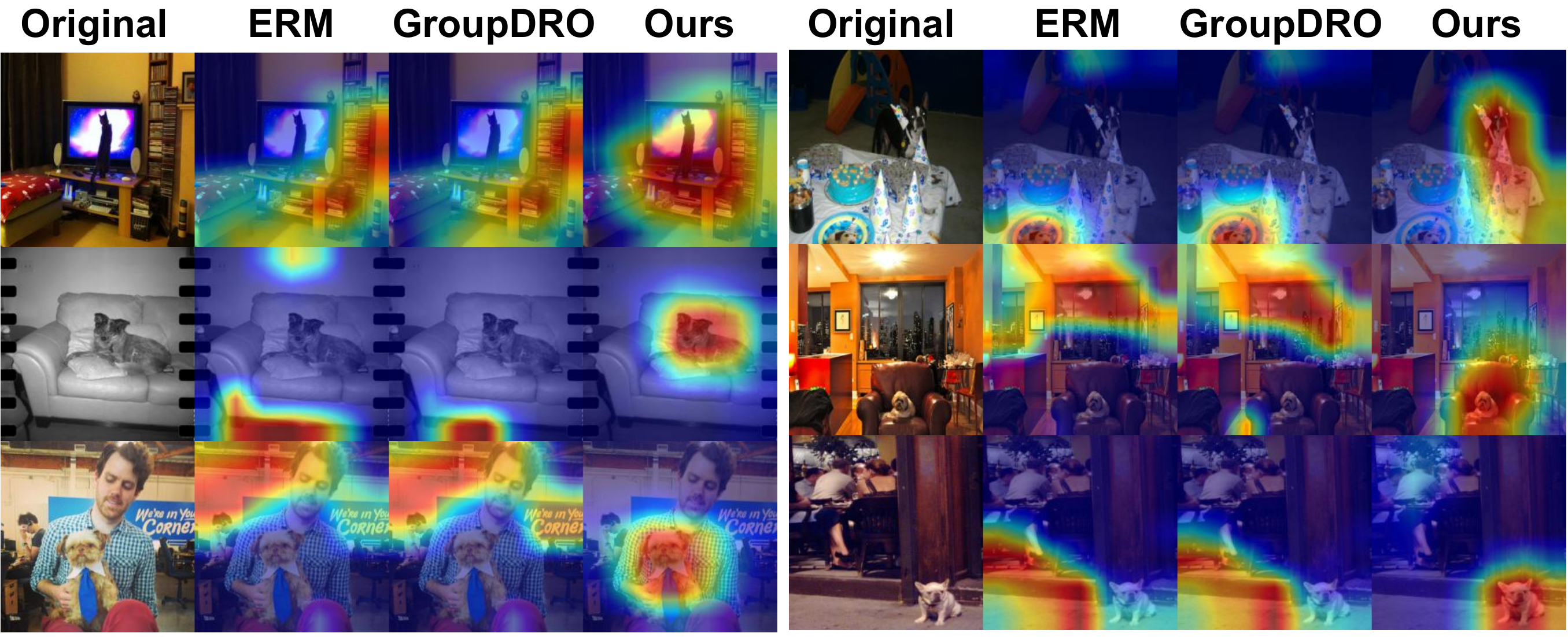}
    \caption{GradCAM of ResNet50 on OOD samples.}
    \vspace*{-3mm}
    \label{fig:gradcam-ood}
\end{figure}

\textbf{Saliency maps.} We plot the saliency maps \citep{selvaraju2017grad} produced when predicting the target attribute of in-domain images in Figure \ref{fig:gradcam-id} and out-of-domain images in Figure \ref{fig:gradcam-ood}. Pixels highlighted by warmer colors indicate higher contributions to the model predictions. Our proposed method can learn causal features rather than focusing on spurious background features when making predictions. Hence, balancing the learning among groups not only de-bias the classifier in-domain but also robustifies it in tackling distributional shifts.
% \vspace*{-3mm}
% \vspace*{-4mm}

\begin{figure}[!ht]
    \centering
     \includegraphics[width=1\columnwidth, ]{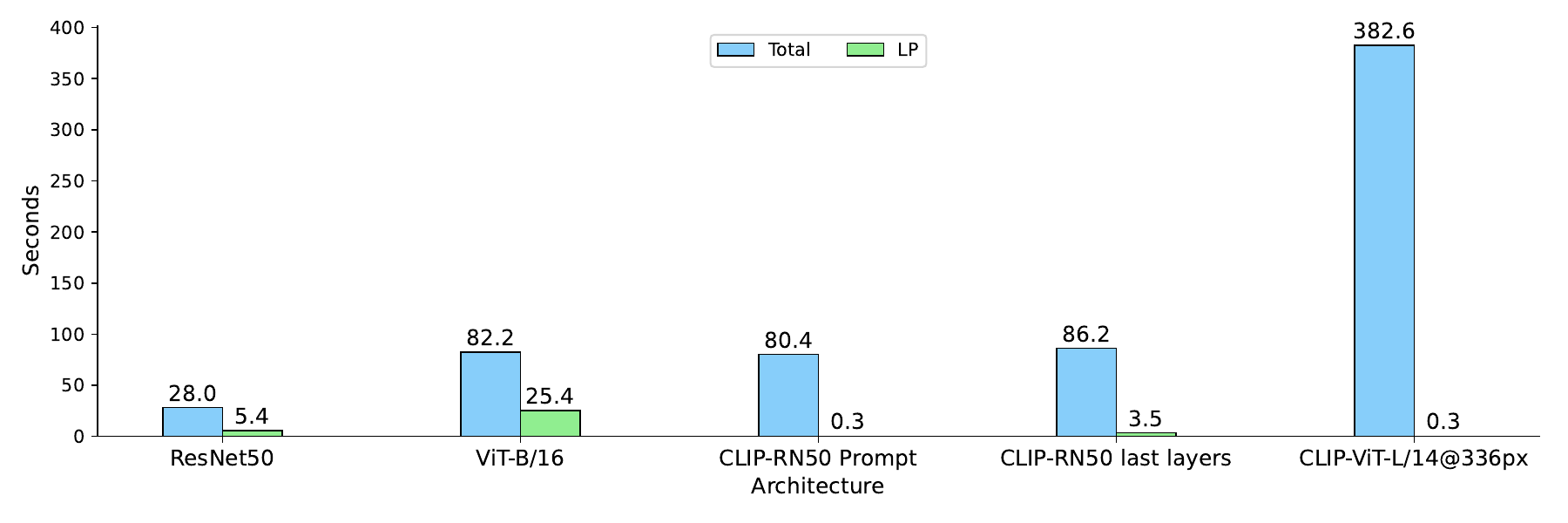}
    \caption{ Total running time and the time for solving the linear programming (LP) problem on epoch with Intel(R) Xeon(R) Platinum 8358 CPU @ 2.60GHz and  NVIDIA A100-SXM4-80GB GPU. Results are averaged over 5 epochs.}
    \vspace*{-2mm}
    \label{fig:runtime}
\end{figure}

\textbf{Running time} Figure \ref{fig:runtime} summarizes the total runtime and the time for solving the linear programming problem on each epoch. The introduction of the optimization (7) in the main paper leads to a small overhead in terms of training time. However, this small sacrifice significantly boosts the performance of different backbones, especially for CLIP and prompt-based models. For example, prompt-tuning on CLIP variants takes only $0.3$ second per epoch since it does not have to update the entire network or even the classification head.

\section{Limitations}
We discuss the limitations of our approach here. While we scale our method when the number of controlling vectors grows via prompt-tuning, we indeed still have to train different prompt sets separately and individually for predetermined values of $c$ (Table \ref{tab:upweight}). Furthermore, in cases where those models could not perform as expected, one might consider incorporating another controlling vector which would require retraining another prompt set from scratch.

To address this limitation, one promising approach is employing hypernetworks \citep{qu2022hmoe,do2023hyperrouter, jiang2023towards}, which take controlling vectors as input to generate corresponding prompt sets. This enables us to train the hypernetwork in an end-to-end fashion without predefining $c$ and maintaining separate prompt sets. Despite this potential limitation, our method is generally tractable and readily applicable.

% Another potential limitation is that the proposed method is somewhat more sophisticated than simple baselines for group robustness, such as deep feature reweighting (DFR), which only requires last-layer retraining. On the other hand, the slight additional complexity of our approach is warranted by superior performance in many settings, for both worst group and average accuracy.

% A further potential limitation is our approach uses group label information. This is a common assumption for group robustness procedures, such as GroupDRO and DFR, but there are recent works that attempt to relax this assumption such as AFR.

\section{Conclusion}
\label{sec:conclusion}
In this paper, we propose Controllable Prompt Tuning, a novel training method that not only improves the performance across groups but also allows us to control the tradeoff between them. Our proposed method can effectively de-bias the training of different types of backbones and achieves state-of-the-art performance on benchmark datasets.
\clearpage
\section*{Acknowledgements}
We thank Chau Pham, Polina Kirichenko, and Shikai Qiu for their helpful discussions.
\section*{Impact Statement}
This paper presents work whose goal is to advance the field of Machine Learning. There are many potential societal consequences of our work, none which we feel must be specifically highlighted here.
\bibliography{ref}
\bibliographystyle{icml2024}

%%%%%%%%%%%%%%%%%%%%%%%%%%%%%%%%%%%%%%%%%%%%%%%%%%%%%%%%%%%%%%%%%%%%%%%%%%%%%%%
%%%%%%%%%%%%%%%%%%%%%%%%%%%%%%%%%%%%%%%%%%%%%%%%%%%%%%%%%%%%%%%%%%%%%%%%%%%%%%%
% APPENDIX
%%%%%%%%%%%%%%%%%%%%%%%%%%%%%%%%%%%%%%%%%%%%%%%%%%%%%%%%%%%%%%%%%%%%%%%%%%%%%%%
%%%%%%%%%%%%%%%%%%%%%%%%%%%%%%%%%%%%%%%%%%%%%%%%%%%%%%%%%%%%%%%%%%%%%%%%%%%%%%

\clearpage
\appendix
\onecolumn

{\begin{center}
 \Large{Supplementary Materials for ``\textbf{Controllable Prompt Tuning for\\ Balancing Group Distributional Robustness}''}   
\end{center}}

Due to space constraints, some details were omitted from the main paper. We therefore include proof for Theorem 4.1 in the main paper (Appendix \ref{sec:app_proof}) and more detailed experimental results (Appendix \ref{sec:app_exp}) in this supplementary material.

\section{Proof of Theorem 4.1}
\label{sec:app_proof}

\textbf{Theorem 4.1. } \textit{Assume that the loss function $\ell$ is differentiable up to the first order with respect to $\theta$, then following}
$$
d_{\text {ent }}:=\sum_{i=1}^K \nabla \ell_i(\theta)\left[p_i \log \left(p_i\right)-p_i \sum_{j=1}^K \log \left(p_j\right) p_j\right]
$$
\textit{where $p_i=\frac{e^{\ell_i(\theta)}}{\sum_{j=1}^K e^{\ell_j(\theta)}}$, maximizes the objective $\mathcal{L}_{\text {ent }}(\theta)$:}
\begin{align*}
   \mathcal{L}_{ent}(\theta) &= H\big(\text{softmax}\big(\vec{\mathcal{L}}(\theta))\big) = H\Big( \text{softmax}\big(\big\{\mathbb{E}_{P_{g^k} }\left[\ell\left(f_\theta(x), y\right)\right] \big\}_{k=1}^K\big)\Big) 
\end{align*}

\textbf{Proof}: For brevity of notations, we denote  $\mathbb{E}_{P_{g^i} }\left[\ell\left(f_\theta(x), y\right)\right]$ as $\ell_i (\theta)$ and $p_i = \frac{e^{\ell_i (\theta)}}{\sum_{j=1}^{K}e^{\ell_j (\theta)}}$. The derivative of $ \mathcal{L}_{ent} $ with respect to $\theta$ is computed as:
\begin{align}
\nabla  \mathcal{L}_{ent}(\theta) &= - \sum_{i=1}^{K} \big\{\nabla p_i [\log (p_i) + 1] \big\}= - \sum_{i=1}^{K} \Bigg\{ [\log (p_i) + 1]  \nabla \frac{e^{\ell_i(\theta)}}{\sum_{j=1}^{K} e^{\ell_j(\theta)}} \Bigg\} \nonumber \\
&=  - \sum_{i=1}^{K} \Bigg\{ [\log (p_i) + 1] \frac{\sum_{j=1}^{K} e^{\ell_j(\theta)} e^{\ell_i(\theta)} \nabla \ell_i(\theta) - e^{\ell_i(\theta)}  \sum_{j=1}^{K} e^{\ell_j(\theta)} \nabla{\ell_j(\theta)} }{(\sum_{j=1}^{K} e^{\ell_j(\theta)})^2} \Bigg\}  \nonumber \\
& =  - \frac{1}{(\sum_{j=1}^{K} e^{\ell_j(\theta)})^2} \sum_{i=1}^{K} \Big\{ [\log (p_i) + 1] e^{\ell_i(\theta)} \sum_{j=1}^{K} e^{\ell_j(\theta)} (\nabla \ell_i(\theta) -\nabla \ell_j(\theta)) \Big\}\nonumber \\
&=    - \frac{1}{(\sum_{j=1}^{K} e^{\ell_j(\theta)})^2} \sum_{i=1}^{K} \Big[  \log (p_i) e^{\ell_i(\theta)} \sum_{j=1}^{K} e^{\ell_j(\theta)} (\nabla \ell_i(\theta) -\nabla \ell_j(\theta))  \Big]\nonumber \\
&=    - \frac{1}{(\sum_{j=1}^{K} e^{\ell_j(\theta)})^2} \sum_{i=1}^{K} \nabla \ell_i(\theta) \Big[ \log (p_i) e^{\ell_i(\theta)} \sum_{j=1}^{K} e^{\ell_j(\theta)} - e^{\ell_i(\theta)} \sum_{j=1}^{K} \log (p_j) e^{\ell_j(\theta)} \Big] \nonumber \\
&=  \sum_{i=1}^{K} \nabla \ell_i(\theta) [p_i \sum_{j=1}^{K} \log (p_j) p_j - p_i\log (p_i) ] 
% \\
% &=   \nabla \mathcal{H}(\theta)   \sum_{i=1}^{K} \Big[ \nabla \ell_i(\theta) - p_i  \nabla \ell_i(\theta) \Big] \nonumber \\
% &\propto \sum_{i=1}^{K}  \Big[ \nabla \ell_i(\theta) - p_i  \nabla \ell_i(\theta) \Big] 
\label{eq:entropy_derivative}
\end{align}

 We conclude the proof. It is noteworthy that when the loss function $\ell$ is positive, one can also simplify the computation by minimizing the following balancing function:
\begin{align*}
 \mathcal{L}'_{ent}(\theta) &:= H\Big(\Big\{  \frac{\mathbb{E}_{P_{g^i} }\left[\ell\left(f_\theta(x), y\right)\right]}{\sum_{j=1}^{K}\mathbb{E}_{P_{g^j} }\left[\ell\left(f_\theta(x), y\right)\right]} \Big\}_{i=1}^K\Big)
\end{align*}
Eliminating the exponential term and denoting $q_i = \frac{{\ell_i (\theta)}}{\sum_{j=1}^{K}{\ell_j (\theta)}}$, the gradient direction of the entropy term is given by:
\begin{align*}
\nabla  \mathcal{L}'_{ent}(\theta) &=  \sum_{i=1}^{K} \big\{\nabla q_i [\log (q_i) + 1] \big\}  
\\  &=  \sum_{i=1}^{K}  \big\{ \frac{{[\sum_{j=1}^{K}{\ell_j (\theta)}] \nabla \ell_i (\theta) - \ell_i (\theta) [\sum_{j=1}^{K}{ \nabla \ell_j (\theta)}] }}{[\sum_{j=1}^{K}{\ell_j (\theta)}]^2} [\log (q_i) + 1] \big\} \nonumber
\\ &\propto \sum_{i=1}^{K} \nabla \ell_i(\theta) \Big\{\log (q_i)   \big[\sum_{j=1}^K \ell_j (\theta) \big] - \sum_{j=1}^K \log ( q_j) \ell_j (\theta) \Big\}
\end{align*}

\section{Experiment details and additional empirical results}
\label{sec:app_exp}

In this section, we first describe the datasets and models used in Appendix \ref{sec:implementation} along with the detailed training configuration for our method, then provide additional results for the experiment in Appendix \ref{sec:addtional_exp}. Example images and data statistics are presented in Table \ref{tab:data_wb}, \ref{tab:data_celeba}, \ref{tab:data_metashift} and \ref{tab:data_isic}.

\subsection{Implementation details}
\label{sec:implementation}

\textbf{Hyper-parameter} We mainly examine the proposed method on ResNet50 \citep{he2016deep}, ViT B/16 \cite{dosovitskiy2020image} and different variants of OpenAI's CLIP\footnote{\url{https://github.com/openai/CLIP}} model.  The hyperparameters selected for each experiment are given in Table \ref{tab:hyperparam}.  Unless stated otherwise, results of baselines are taken from original papers and \citep{deng2023robust, wu2023discover, piratla2021focus, zhang2022correct, kim2023removing}, which provide standard evaluation protocols for different backbones and datasets. Performance of ERM and GroupDRO \citep{sagawa2019distributionally} at different amount of training data and their training curves are obtained from the released codebase of GroupDRO\footnote{\url{https://github.com/kohpangwei/group_DRO}} using their provided commands.

\begin{table}[!ht]
\centering
  \resizebox{0.8\textwidth}{!}{%
\begin{tabular}{lccccccc}
\toprule Dataset  & Architecture & Learning Rate & Weight Decay & Batch Size & \# Epochs & Prompt length \\
\midrule
\midrule
Waterbirds & ViT B/16   & 0.1 & 0.001 & 64 & 100 & 10 \\
Waterbirds &  Last layers CLIP-RN50  & 0.003 & 0.01 & 32 & 100 & - \\
Waterbirds &  Prompt CLIP-RN50 & 0.03 & 0.01 & 32 & 100 & 16 \\
Waterbirds &  Last layer CLIP-ViT-L/14  & 0.01 & 0.01 & 32 & 100 & - \\
Waterbirds &  Prompt CLIP-ViT-L/14  & 0.001 & 0.01 & 32 & 100 & 16 \\
CelebA &  ViT B/16   & 0.01 & 0.001 & 128 & 100 & 10 \\
MetaShift &  ResNet50  & 0.003 & 0.01 & 16 & 100 & - \\
MetaShift &  ViT B/16  & 0.01 & 0.1 & 16 & 100 & 5 \\
ISIC &  ViT B/16  & 0.003 & 0.01 & 16 & 100 & 5 \\

\bottomrule
\end{tabular}}
  \caption{Hyperparameter for different experiments throughout our paper. We report the hyper-parameters selected for our proposed method after performing grid-search. \label{tab:hyperparam}}
\end{table}

\textbf{Classifier design.} While we train a classification head from scratch for ResNet50 or ViT, CLIP allows us to utilize its meaningful multimodality embeddings as a zero-shot classifier. In particular, given an image embedding $\mathbf{e}_i$ and a text class embedding $\mathbf{e}^k_t$ for some class index $k \in\{1,2, \ldots, \mathbf{N_c}\}$, where $\mathbf{N}_c$ is the number of categories. The probability of the given image belonging to $k$-th class is computed as:
$$
p(y=k \mid \mathbf{e}_i)=\frac{\exp \left(<\mathbf{e}_i, \mathbf{e}^k_t>/ \tau\right)}{\sum_{j=1}^{\mathbf{N}_c} \exp \left(< \mathbf{e}_i, \mathbf{e}^j_t>/ \tau\right)}
$$
where $<\cdot, \cdot>$ denotes the cosine similarity and $\tau$ is the temperature parameter.

\textbf{Dataset statistics.} We  evaluate CPT on  Waterbirds \citep{sagawa2019distributionally} (Table \ref{tab:data_wb}), CelebA \citep{liu2015faceattributes} (Table \ref{tab:data_celeba}), MetaShift \citep{liang2021metashift} (Table \ref{tab:data_metashift}) and ISIC \citep{codella2019skin} (Table \ref{tab:data_isic}), following previous work \cite{sagawa2019distributionally, deng2023robust, wu2023discover}. While the objective of those image classification tasks is predicting the categories of input images, those target attributes are often correlated well with spurious features (please refer to each table for detailed descriptions of spurious features and group partitions for each dataset). 

\begin{table}[!ht]
\centering
  \resizebox{0.7\textwidth}{!}{%
  \begin{tabular}{l|cccc}

\toprule {   Image}
& \includegraphics[width=2.6cm, height=1.8cm]{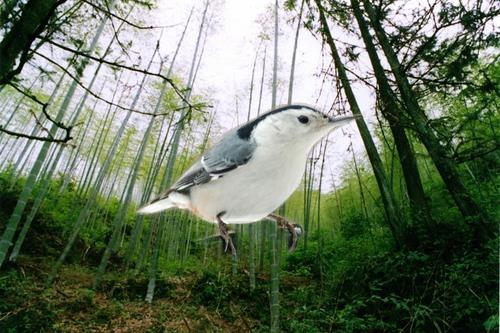} 
& \includegraphics[width=2.6cm, height=1.8cm ]{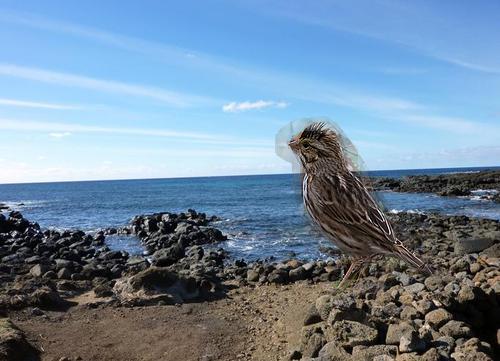} 
& \includegraphics[width=2.6cm, height=1.8cm]{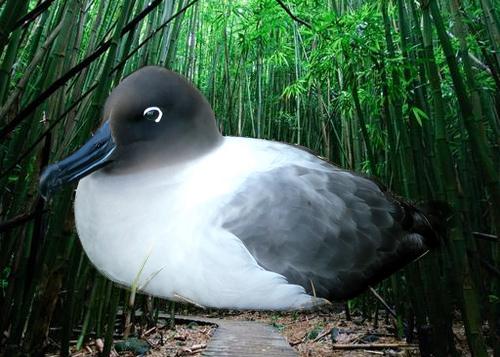} 
& \includegraphics[width=2.6cm, height=1.8cm]{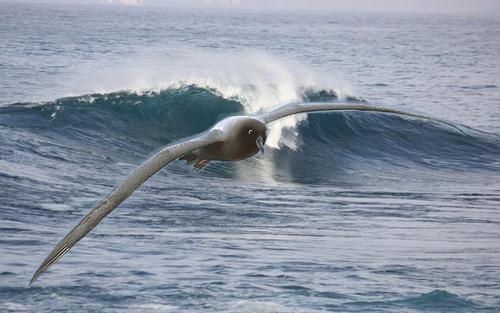} 
\\
\midrule
 Group   & 0 & 1 & 2 & 3 \\
Label & 0  (landbird) & 0  (landbird) & 1  (waterbird) & 1   (waterbird) \\
 Spurious feature   & 0 (land) & 1 (water) & 0 (land) & 1 (water) \\
 Description & landbird on land & landbird on water & waterbird  on land & waterbird  on water \\
 \# Train data & $3,498 (73 \%)$ & $184 (4 \%)$ & $56 (1 \%)$ & $1,057 (22 \%)$ \\
 \# Val data & $467$ & $466$ & $133$ & $133$ \\
\# Test data & $2,255$ & $2,255$ & $642$ & $642$ \\
\bottomrule
\end{tabular}}
  \caption{Example images of Waterbirds \citep{sagawa2019distributionally}. \label{tab:data_wb}}
\end{table}

\begin{table}[!ht]
\centering
  \resizebox{0.7\textwidth}{!}{%
  \begin{tabular}{l|cccc}

\toprule {   Image}
& \includegraphics[width=2.6cm, height=2.6cm]{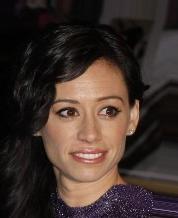} 
& \includegraphics[width=2.6cm, height=2.6cm ]{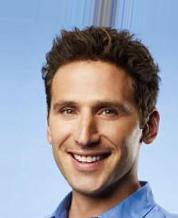} 
& \includegraphics[width=2.6cm, height=2.6cm]{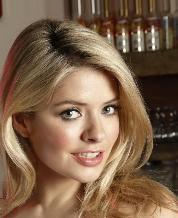} 
& \includegraphics[width=2.6cm, height=2.6cm]{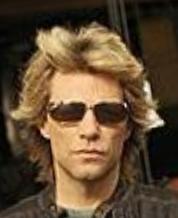} 
\\
\midrule
 Group   & 0 & 1 & 2 & 3 \\
Label & 0  (non-blond) & 0  (non-blond) & 1  (blond) & 1   (blond) \\
 Spurious feature   & 0 (woman) & 1 (man)  & 0 (woman) & 1 (man) \\
 Description & non-blond woman & non-blond man & blond woman & blond man \\
 \# Train data & $71,629 (44 \%)$ & $66,874 (41 \%)$ & $22,880 (14 \%)$ & $1,387 (1 \%)$ \\
 \# Val data & $8,535$ & $8,276$ & $2,874$ & $182$ \\
\# Test data & $9,767$ & $7,535$ & $2,480$ & $180$ \\
\bottomrule
\end{tabular}}
  \caption{Example images of  CelebA \citep{liu2015faceattributes}. \label{tab:data_celeba}}
\end{table}

\begin{table}[!ht]
\centering
  \resizebox{0.85\textwidth}{!}{%
\begin{tabular}{l|cccc|cc}
\toprule

Image
& \includegraphics[width=2.6cm, height=1.8cm]{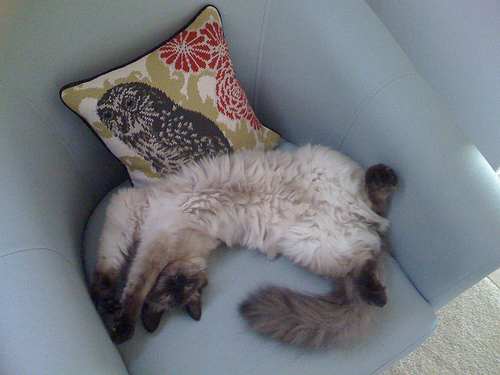} 
& \includegraphics[width=2.6cm, height=1.8cm ]{fig/examples/cat_bed2.jpg} 
& \includegraphics[width=2.6cm, height=1.8cm]{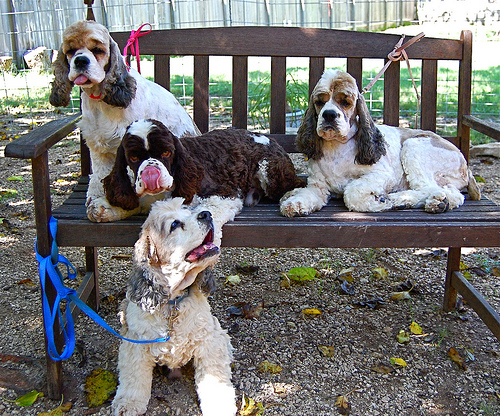} 
& \includegraphics[width=2.6cm, height=1.8cm]{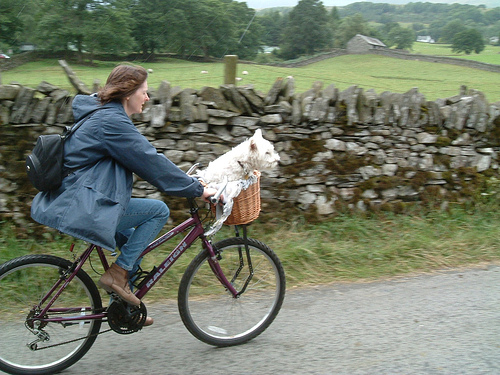} 
& \includegraphics[width=2.6cm, height=1.8cm]{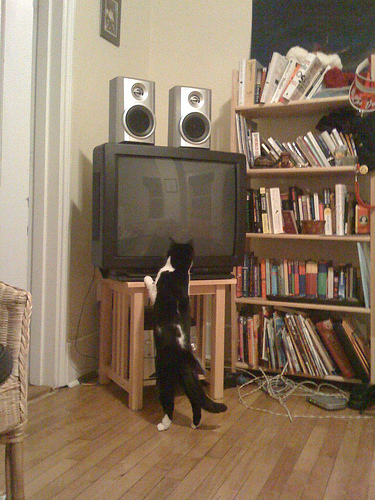} 
& \includegraphics[width=2.6cm, height=1.8cm]{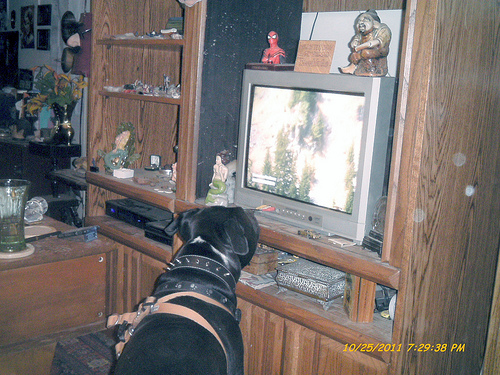} 
\\
\midrule
 Group  & 0 & 1 & 2 & 3 & 4 & 5 \\
Label & 0 (cat) & 0 (cat) & $1$(dog) & $1$ ({dog}) & 0 (cat) & $1$ ({dog})\\
Spurious feature & 0 (sofa) & 1 (bed) & 2 (bench) & 3 (bike) & 4 (shelf) & 4 (shelf) \\
\# Train data & 231 & 380 & 145 & 367 & - & - \\
\# Val data (OOD) & - & - & - & - & 34 & 47 \\
\# Test data & - & - & - & - & 201 & 259 \\
\bottomrule
\end{tabular}}
  \caption{Example images of  MetaShift \citep{liang2021metashift}. \label{tab:data_metashift}}
\end{table}

\begin{table}[!ht]
\centering
  \resizebox{0.85\textwidth}{!}{%
\begin{tabular}{c|cccccc}
\toprule
Image
& \includegraphics[width=2.6cm, height=1.8cm]{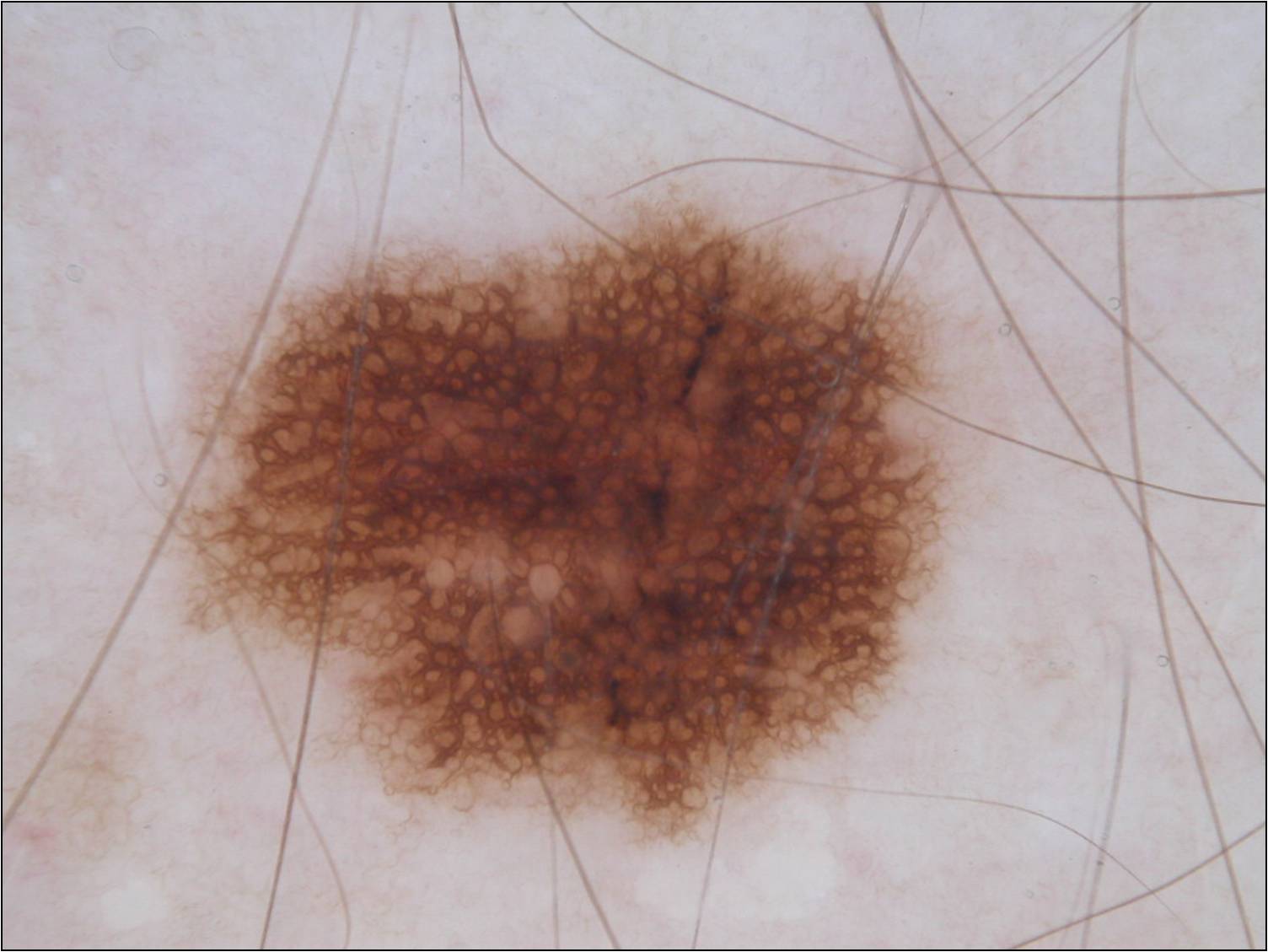} 
& \includegraphics[width=2.6cm, height=1.8cm ]{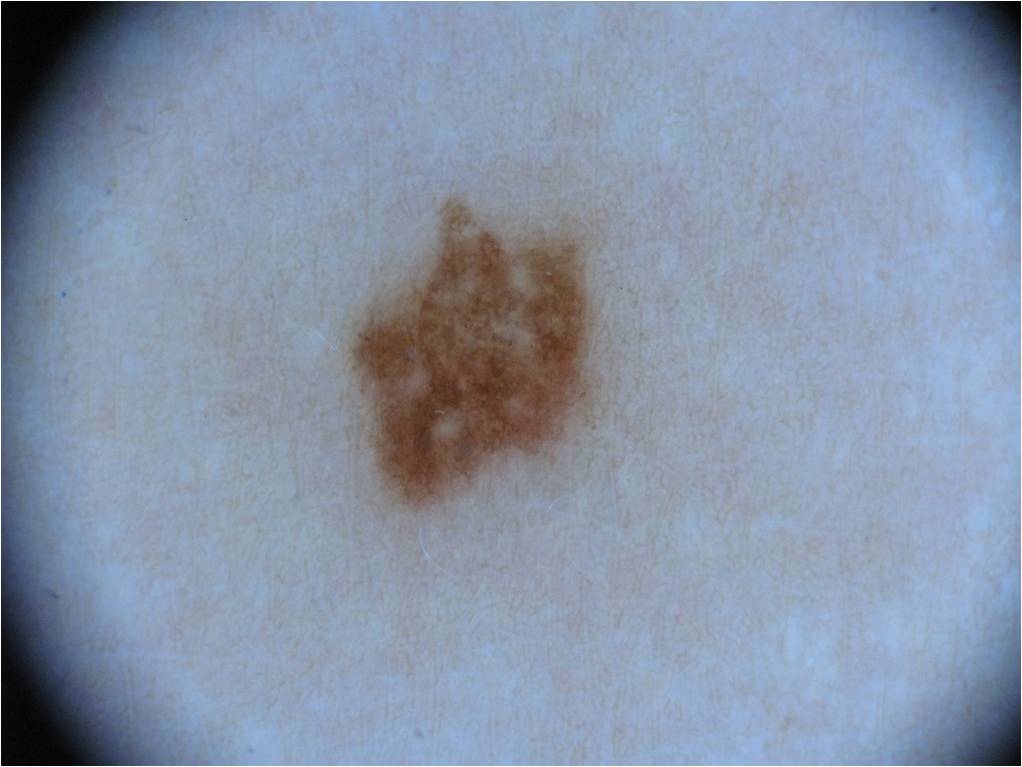} 
& \includegraphics[width=2.6cm, height=1.8cm]{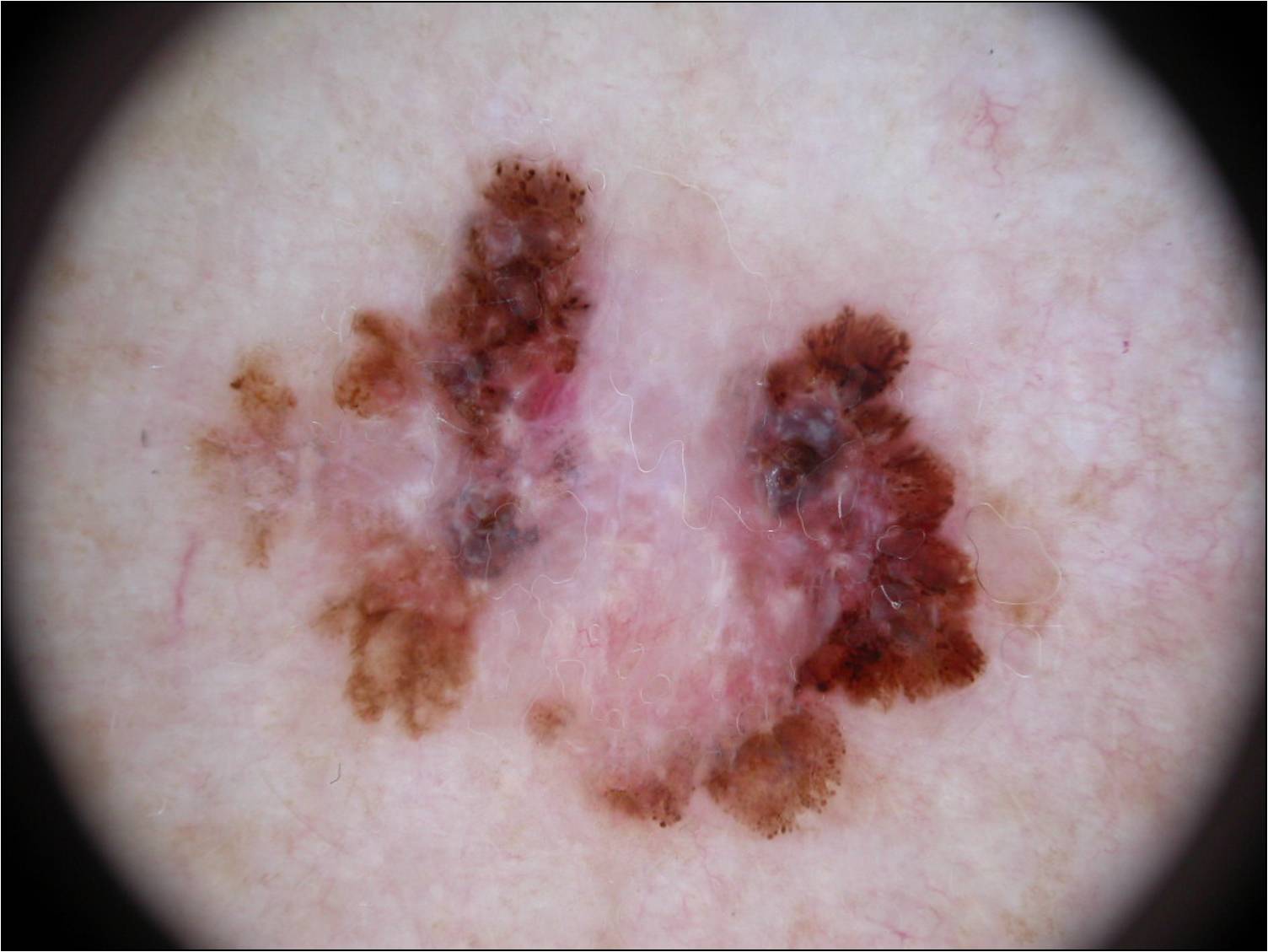} 
& $\ldots$ 
& \includegraphics[width=2.6cm, height=1.8cm]{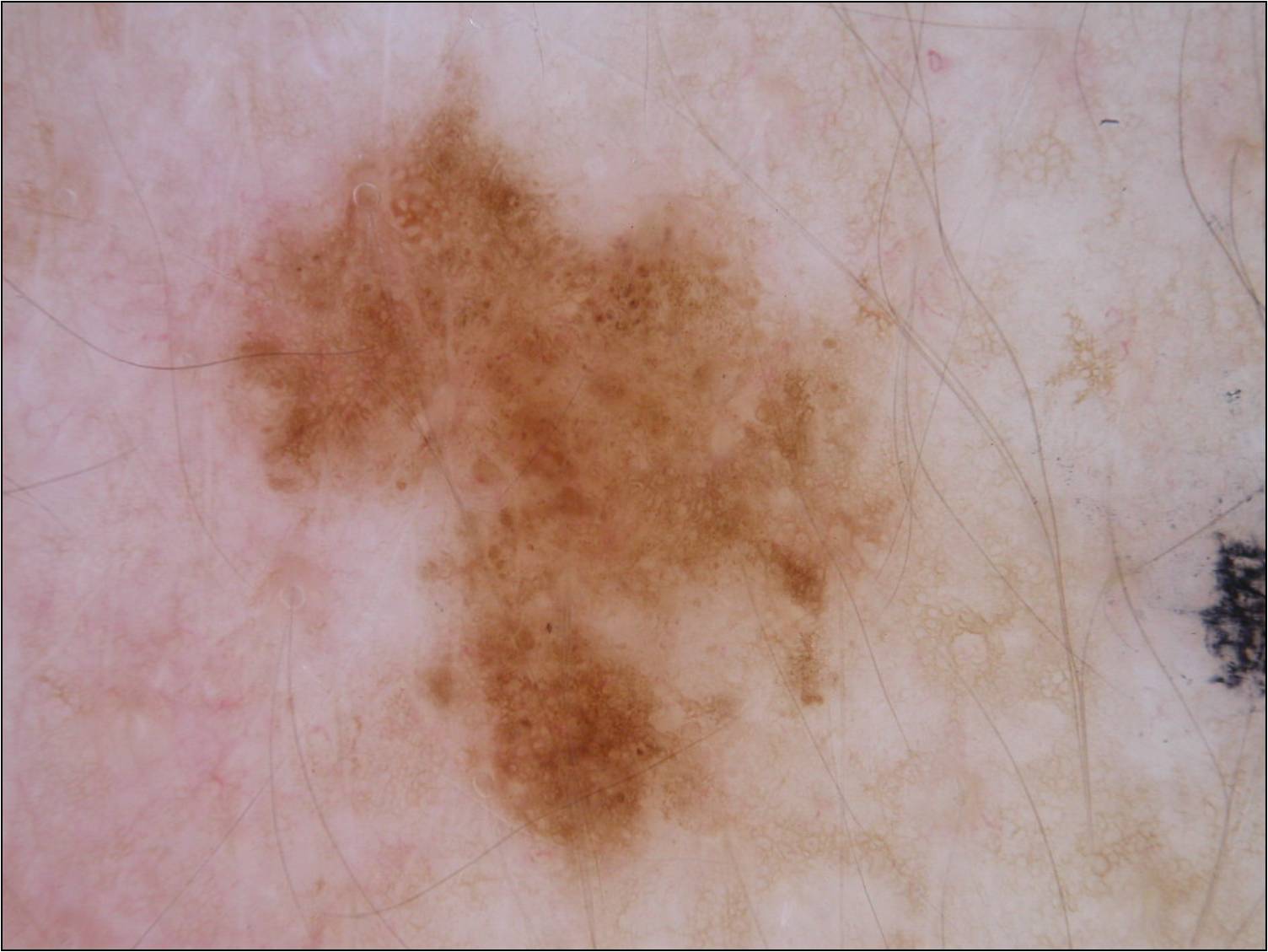} 
& \includegraphics[width=2.6cm, height=1.8cm]{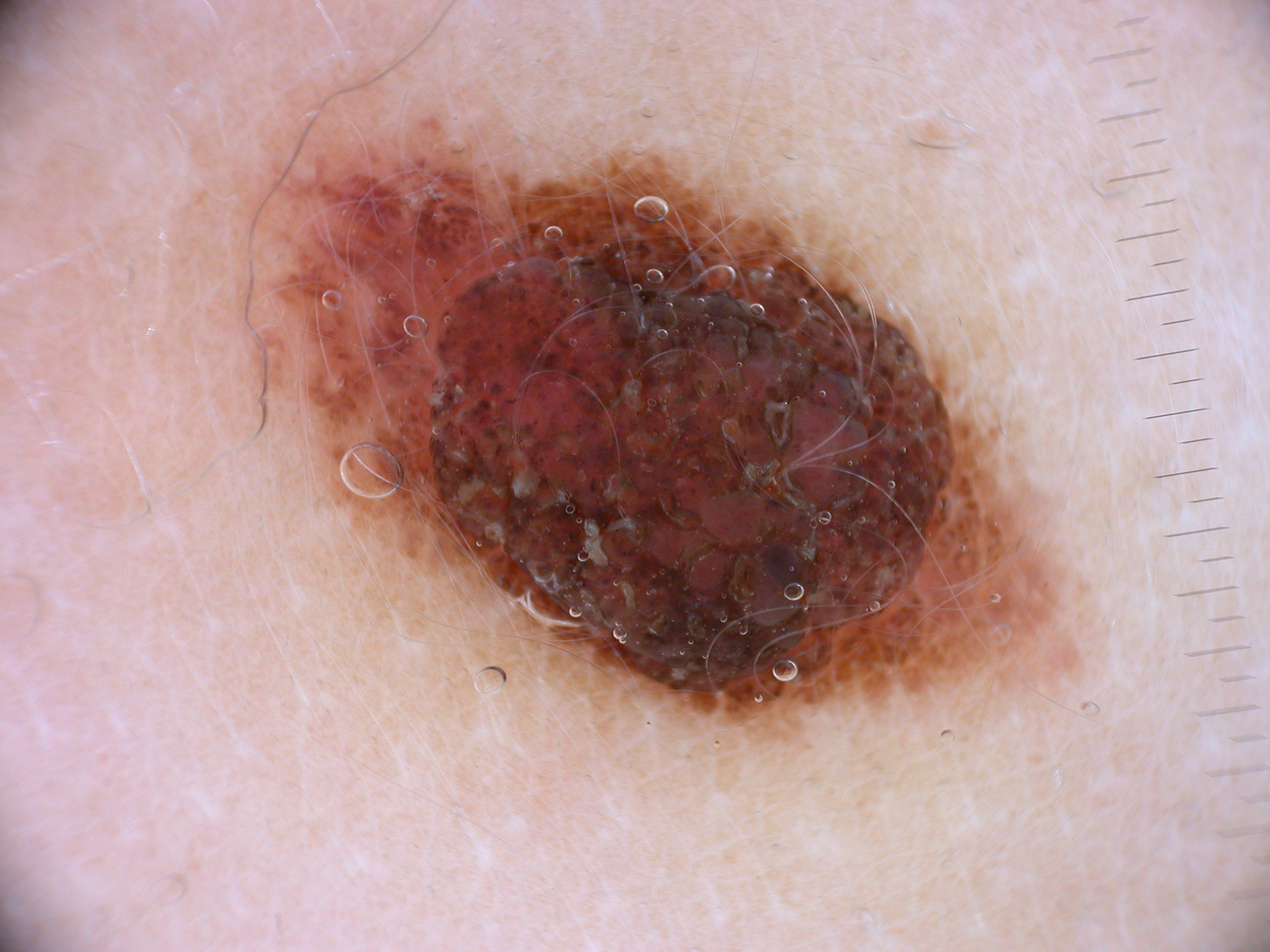} 
\\
\midrule
Label & 0 (benign) & 0 (benign)& 1 (malignant) & $\ldots$ & 0 (benign) & 1 (malignant) \\
 Spurious feature & hair& dark corner& gel bubbles & $\ldots$ & ink & ruler \\
 \# Train data: & \multicolumn{6}{c}{1,826} \\
 \# Val data: & \multicolumn{6}{c}{154} \\
 \# Test data: & \multicolumn{6}{c}{618} \\
\bottomrule
\end{tabular}}
  \caption{Example images of  ISIC \citep{codella2019skin}. \label{tab:data_isic}}
\end{table}

\clearpage
\subsection{Additional experiments}
\label{sec:addtional_exp}

\textbf{Prompt length}  Thanks to the design of the prompting component, the computational burden during training is modest: only a small portion of the model parameters are updated through back-propagation. Here, we vary the value of prompt length $L$ to observe its effect on the model performance in Table \ref{tab:prompt_length}. Results of ERM, GroupDRO, the second-best method in this dataset DFR \cite{kirichenko2022last} and a recent baseline in this line of research PDE \cite{deng2023robust} are also included for a general comparison. While increasing the prompt length can improve the performance of the model, we find that just pretending one token per transformer layer ($L=1$) can outperform the second-best method that requires $\times2200$ more trainable parameters via better training. Last, it is noteworthy to highlight that $L=16$ can help CPT match the average accuracy score of ERM in this scenario, which demonstrates the large modeling capacity of our proposed method.

\begin{table}[!ht]
  \caption{Ablation results for different prompt length $N$ on ViT B/16 on Waterbirds.}
  \label{tab:prompt_length}
  \centering
  \resizebox{1\textwidth}{!}{%
\begin{tabular}{lcccccc|cccc}
\toprule
$N$ & $1$ & $2$   & $4$    & $8$  & $10$  & $16$ & ERM & GroupDRO & PDE & DFR\\
\midrule Worst & $92.9_{\pm 0.8} $ & $93.3_{\pm 0.9} $  &  $93.5_{\pm 0.6} $  &  $93.8_{\pm 0.5} $ &  $93.5_{\pm 0.4} $ &  $\mathbf{93.9}_{\pm 0.7} $ &  $70.0_{\pm 2.3} $ &  $86.7_{\pm 0.6} $  &  $90.3_{\pm 0.3} $  &  $92.9_{\pm 0.2} $  \\
Average  &  $95.8_{\pm 0.9} $  &  $95.9_{\pm 0.9} $  &  $96.4_{\pm 0.1} $ &  $96.5_{\pm 0.0} $ &  $96.3_{\pm 0.1} $ &  $\mathbf{97.1}_{\pm0.3} $ &  $\mathbf{97.1}_{\pm 0.1} $ &  $93.2_{\pm 0.5} $ &  $92.4_{\pm 0.8} $ &  $94.2_{\pm 0.2} $\\
Mean  &  $94.9_{\pm 0.8} $ &  $95.2_{\pm 0.0} $ &  $95.6_{\pm 0.4} $ &  $\mathbf{96.0}_{\pm 0.8} $ &  $95.7_{\pm 0.2} $ &  $95.7_{\pm 0.3} $ &  N/A  &  N/A   &  N/A  &  N/A \\
\# Params & 10754 &  19970 &  38402 & 75266 & 93698 &  148994 & 23512130 & 23512130 & 23512130 & 23512130 \\
\bottomrule
\end{tabular}}
\end{table}

\begin{figure*}[!ht]
    \centering
     \includegraphics[width=1\columnwidth, ]{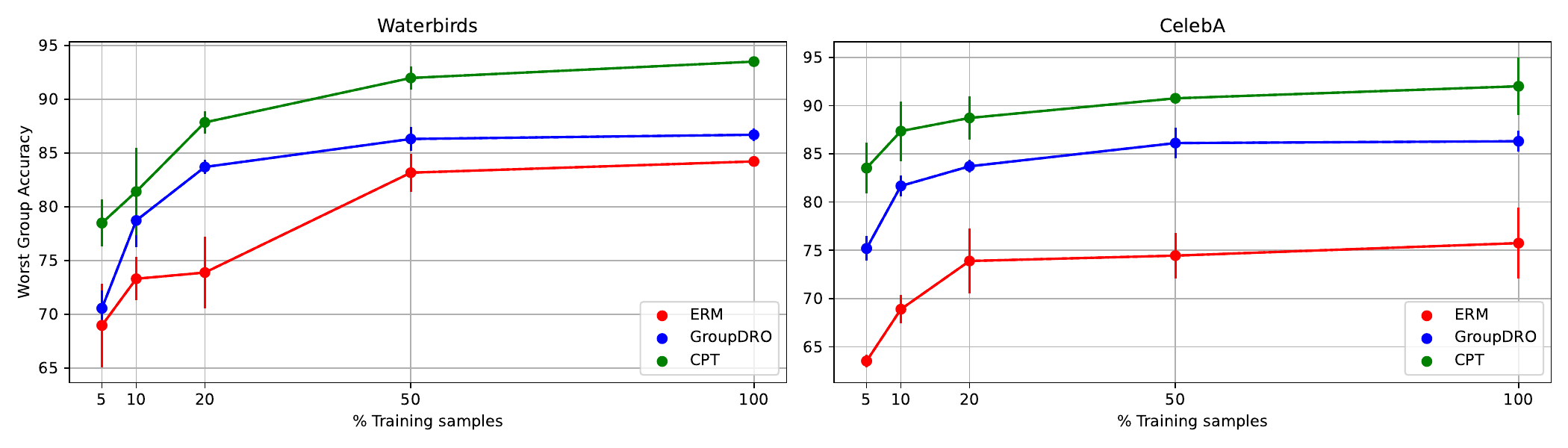}
    \caption{Data Efficiency on Waterbirds (left) and CelebA (right). Error bars correspond to standard deviations over three trials.}
    % \vspace*{-2mm}
    \label{fig:wb_celebA_data_efficiency}
\end{figure*}

\textbf{Performance at different number of training samples.} In order to comprehensively assess the performance of our proposed model, we conduct a series of experiments to analyze its behavior across varying sizes of training data. Those experiments were designed to investigate the impact of data size on the model performance of comparative methods.  We keep group ratios preserved while subsampling a part of the training data. The x-axis of Figure \ref{fig:wb_celebA_data_efficiency} denotes the percentages of the Waterbirds and CelebA training set used in each sub-experiment. It can be seen that CPT is better than GroupDRO or ERM in lower data regimes. Remarkably, by using $20\%$ and $10\%$ data, our proposed method can still outperform other baselines when using the full training set.

\begin{figure*}[!ht]
    \centering
     \includegraphics[width=.95\columnwidth, ]{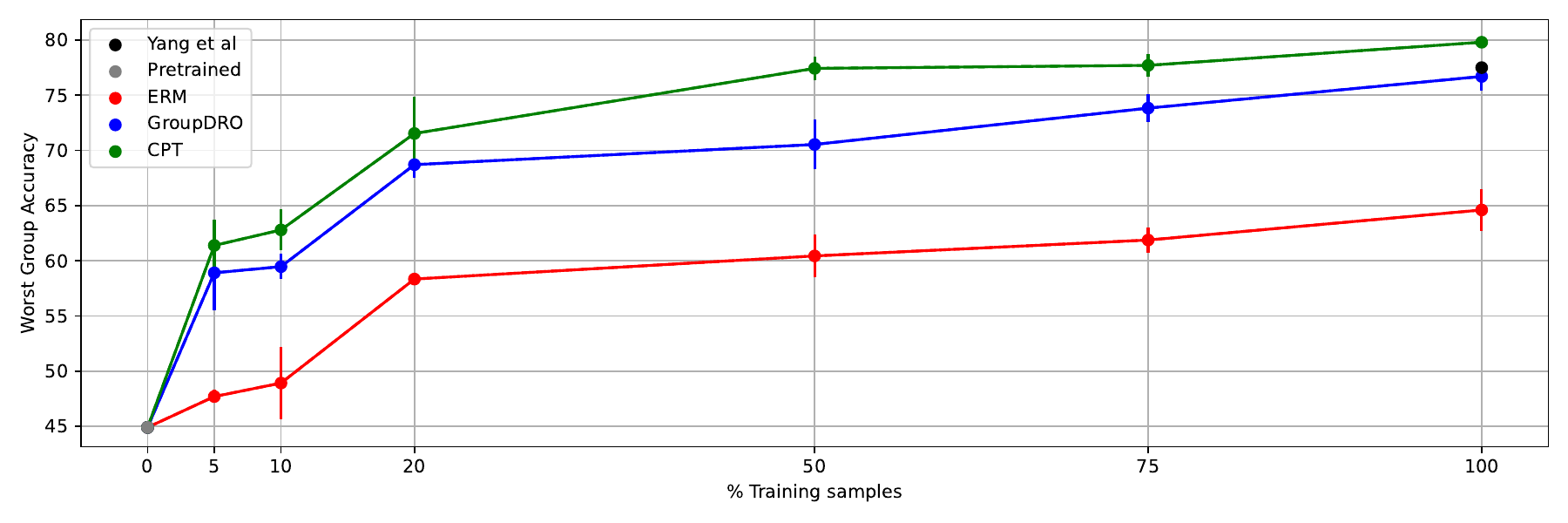}
    \caption{Data Efficiency of CLIP-RN50.  CPT with $75\%$ data can outperform the current state-of-the-art method for debiasing CLIP \cite{yang2023mitigating} which uses $\times 1800$ trainable parameters.
}
    % \vspace*{-2mm}
    \label{fig:clip_data_efficiency}
\end{figure*}

Similarly, we run the above experiment but replace the main backbone with CLIP-RN50 and plot the performance of different baselines. Note that we freeze both image and text encoders and employ prompt tuning on their text encoder only. The text prompt used for training CLIP models is:
$$ \textit{``a type of bird, a photo of a waterbird/landbird''}$$
While the pretrained CLIP model obtains $90.9\%$ average accuracy on Waterbirds (Table 2 in the main paper), it severely suffers from unintended bias when attaining only $44.9\%$ worst-group accuracy. This motivates the need for debiasing methods to remove such bias and mitigate the reliance on spurious features when making predictions of foundation models. Figure \ref{fig:clip_data_efficiency} shows that while using more data can help enhance the pre-trained CLIP model better for all methods, CPT obtains the best worst group accuracy for every amount of data and surpasses the current state-of-the-art method \cite{yang2023mitigating}.

\clearpage

\begin{wraptable}{r}{0.68\textwidth}
\vspace*{-5mm}

\caption{Comparison of different backbones on Waterbirds \label{tab:vit_variant}}
\centering
\resizebox{.65\columnwidth}{!}{\begin{tabular}{lccc}

\hline Model   & Average Acc. & Worst-Group Acc.  & \# Params\\
\midrule
BiT-M-R50x1  & $92.05 _{\pm 0.05}$ & $75.10 _{\pm 0.62}$ & 23.5M \\
BiT-M-R50x3  & $94.90 _{\pm 0.05}$ & $80.51 _{\pm 1.02}$ & 211M\\
BiT-M-R101x1  & $94.05 _{\pm 0.07}$ & $77.50 _{\pm 0.50}$ & 42.5M \\
ViT-Ti/16 & $89.50 _{\pm 0.05}$ & $71.65 _{\pm 0.16}$ & 5.6M \\
ViT-S/16 & $96.30 _{\pm 0.51}$ & $85.45 _{\pm 1.16}$ & 21.8M\\
ViT-B/16 & ${9 6 . 7 5}_{ \pm {0 . 0 5}}$ & ${8 9 . 3 0} _{\pm {1 . 9 5}}$ & 86.1M \\
\midrule
CPT  & ${{96.33}}_{\pm
0.15}$  & ${\mathbf{93.51}}_{\pm 0.36} $ & $\mathbf{94}$k \\
\hline
\end{tabular}
}
\end{wraptable}
\textbf{Debiasing ViT.} For the experiment of debiasing Vision Transformer, we compare CPT against different ViT and Big Transfer (BiT) models introduced in \cite{kolesnikov2020big, ghosal2023vision}. More detailed results are given in Table \ref{tab:vit_variant}, from which we can see that CPT not only obtains the best worst-group accuracy score but also performs on par with the second-best method in terms of average accuracy while using much fewer parameters. Note that other models scale up the resolution to $384 \times 384$ while CPT keeps using the $224\times224$ resolution.

\textbf{Performance on CLIP variants.}  Table \ref{tab:clip_variant} presents the scores of ERM, GroupDRO, and CPT on a wide range of CLIP backbones, including CLIP with ResNet101, ViT-B/16, ViT-B/32, ViT-L/14. Here we simply use the balanced version of CPT instead of hyperparameter tuning on $c$. Therefore, note that we do not try to beat GroupDRO on all metrics, but focus on investigating how our optimization procedure helps close the gap among groups. In summary, balancing the learning among groups results in large improvements in worst-group and mean accuracies for all backbones.

\begin{table*}[!ht]
%\vspace*{-2mm}
\caption{Performance on other CLIP variants, where CPT has the lowest gap between worst and average accuracy.\label{tab:clip_variant}}%\vspace*{-2mm}
  \centering
  \resizebox{.95\textwidth}{!}{\begin{tabular}{l|ccc|ccc|ccc|ccc}
\midrule Model & \multicolumn{3}{c|}{ Last layers ResNet101} & \multicolumn{3}{c|}{ Prompting ResNet101} & \multicolumn{3}{c|}{ Last layer ViT-B/16} & \multicolumn{3}{c|}{ Prompting ViT-B/16 }  \\
 \# Params & \multicolumn{3}{c|}{13740544} & \multicolumn{3}{c|}{8192}&  \multicolumn{3}{c|}{393216} & \multicolumn{3}{c}{8192} \\
Accuracy & Average & Worst & Mean  & Average & Worst & Mean & Average & Worst & Mean & Average & Worst & Mean  \\
\midrule Pre-trained  & $91.7$ & $46.8$ & $71.1$ &  $91.7$ & $46.8$ & $71.1$    & $90.6$  & $50.1$ & $75.5$   & $90.6$  & $50.1$ & $75.5$   \\
ERM & $89.3$ & $69.4$ & $80.2$ & $88.6$ & $67.2$ & $80.0$ & $95.7$ & $77.4$ & $87.8$  & $96.1$ & $82.2$ & $89.7$  \\
GroupDRO & $88.5$ & $74.9$ & $82.1$ & $86.3$ & $74.0$ & $80.7$  & $91.4$ & $88.5$ & $90.2$  & $93.3$ & $85.5$ & $90.4$ \\ 
CPT & $87.7$ & $\mathbf{75.2}$ & $\mathbf{83.9}$ & $83.1$ & $\mathbf{78.9}$ & $\mathbf{81.2}$ & $91.4$  & $\mathbf{89.4}$ & $\mathbf{90.5}$  & $92.2$ & $\mathbf{89.2}$ & $\mathbf{91.5}$ \\ 
\midrule
\midrule
 Model & \multicolumn{3}{c|}{ Last layer ViT-B/32} & \multicolumn{3}{c|}{ Prompting ViT-B/32 }  & \multicolumn{3}{c|}{ Last layer ViT-L/14 } & \multicolumn{3}{c}{ Prompting ViT-L/14 }   \\
 \# Params & \multicolumn{3}{c|}{393216} & \multicolumn{3}{c|}{8192}& \multicolumn{3}{c|}{786432}& \multicolumn{3}{c}{12288} \\
Accuracy & Average & Worst & Mean  & Average & Worst & Mean & Average & Worst & Mean & Average & Worst & Mean  \\
\midrule
Pretrained & $90.1$ & $49.0$ & $72.2$  & $90.1$ & $49.0$ & $72.2$ & $89.4$  & $36.1$ & $83.9$ & $89.4$  & $36.1$ & $83.9$ \\
ERM & $94.7$ & $73.5$ & $85.2$ & $94.0$ & $76.0$ & $85.7$ &  $96.6$  & $85.8$ & $91.8$ & $96.8$ & $83.1$ & $91.6$ \\
GroupDRO & $90.7$ & $81.6$ & $86.8$ & $88.3$ & $82.3$  & $85.9$ & $95.8$ & $88.9$  & $92.8$ & $95.5$ & $86.3$ & $91.5$ \\
CPT & $87.4$ & $\mathbf{86.3}$ & $\mathbf{87.4}$ & $86.8$ & $\mathbf{85.2}$  & $\mathbf{87.5}$ & $94.8$ & $\mathbf{90.7}$ & $\mathbf{92.9}$ & $94.9$ & $\mathbf{90.0}$ & $\mathbf{92.6}$ \\
\bottomrule
\end{tabular}}
%\vspace*{-4mm}
\end{table*}

\textbf{Prompt-tuning with decoupled backbones.} While the proposed method works well with CLIP, which has jointly-trained image and text encoders, we also examine it on decoupled backbones to show that the efficiency of CPT does not rely on powerful pretrained models. We thus accordingly expanded our experimental scope to include decoupled backbones and provide the results in Table \ref{tab:decoupled}. Since there is some difference between the original architecture of CLIP's image encoder and the ImageNet-pre-trained encoder, we need to train additional bottleneck layers to map both image and text inputs to a unified dimensional space.

\begin{table}[!ht]

\caption{Performance of decoupled vision-language backbones on Waterbirds \label{tab:decoupled}}
\centering
\resizebox{.6\columnwidth}{!}{\begin{tabular}{c|c|c|c|c}
\toprule
 Backbone & Method & Average & Worst & \# Params \\
 \midrule
\multirow[t]{4}{*}{ CLIP-RN50 } & Pre-trained & 90.8 & 44.9 & 0 \\
 & ERM & 85.7 & 64.6 & 8192 \\
 & GroupDRO & 81.0 & 76.7 & 8192 \\
 & Ours & 81.8 & 79.8 & 8192 \\
 \midrule
\multirow[t]{3}{*}{ CLIP \& RN50 ImageNet } & ERM & 93.9 & 69.8 & 2105344 \\
& GroupDRO & 92.3 & 75.7 & 2105344 \\
 & Ours & 92.4 & 79.3 & 2105344 \\
 \midrule
 \multirow[t]{4}{*}{ CLIP-ViT-B/16 } & Pre-trained & 90.6 & 50.1 & 0 \\
 & ERM & 96.1 & 82.2 & 8192 \\
 & GroupDRO & 93.3 & 85.5 & 8192 \\
 & Ours & 92.2 & 89.2 & 8192 \\
 \midrule
 \multirow[t]{3}{*}{ CLIP \& ViT-B/16 ImageNet } & ERM & 75.5 & 47.9 & 401408 \\
 & GroupDRO & 71.5 & 55.5 & 401408 \\
& Ours & 70.9 & 58.6 & 401408 \\
\bottomrule
\end{tabular}}
\end{table}

Since there are mismatches between how the image and text encoders comprehend and represent information, the performance of those decoupled CLIP models drop significantly. However, in such cases, CPT still consistently exhibits superior worst-group accuracy compared to ERM or GroupDRO.

\textbf{Label noise robustness.} CPT is empowered by prompt-tuning, which has been shown to be more robust to noisy labels than full fine-tuning or linear probing, as discussed in \citep{wu2023prompt}, underscoring its practical advantage in the outliers problem. To examine the robustness against noisy labels of different training methods, we conducted an additional experiment on the CLIP-RN50 and Waterbird dataset under different levels of label noise, ranging from $20\%$ to $60\%$.

\begin{table}[!ht]
\caption{Accuracy for models trained on Waterbirds with different levels of noisy labels. \label{tab:noisy_label}}
\centering
\resizebox{1\columnwidth}{!}{
\begin{tabular}{l|ccc|ccc|ccc} 
\toprule
& &{$\mathbf{2 0 \%}$}  & & & $\mathbf{4 0 \%}$ & & & $\mathbf{6 0 \%}$ \\
\midrule Method & Average & Mean & Worst & Average & Mean & Worst & Average & Mean & Worst \\
\toprule
 Last Layers \citep{yang2023mitigating} & $79.27 \pm 2.74$ & $75.85 \pm 1.89$ & $66.15 \pm 2.31$ & $67.16 \pm 2.20$ & $63.70 \pm 2.32$ & $57.84 \pm 2.12$ & $42.51 \pm 1.49$ & $45.59 \pm 3.72$ & $39.37 \pm 1.50$ \\
 Linear Probing & $79.47 \pm 2.48$ & $75.76 \pm 0.42$ & $70.44 \pm 2.67$ & $67.05 \pm 2.56$ & $63.15 \pm 0.96$ & $57.06 \pm 1.61$ & $40.16 \pm 1.48$ & $40.79 \pm 2.05$ & $35.39 \pm 2.08$ \\
 Prompt-Tuning & $77.66 \pm 0.61$ & $74.20 \pm 1.43$ & $70.35 \pm 2.17$ & $69.76 \pm 2.89$ & $64.71 \pm 1.33$ & $58.82 \pm 1.67$ & $50.00 \pm 1.14$ & $46.69 \pm 1.60$ & $42.76 \pm 1.73$ \\
 CPT & $76.57 \pm 1.65$ & $75.38 \pm 1.69$ & $75.14 \pm 2.51$ & $67.67 \pm 2.02$ & $67.48 \pm 0.43$ & $61.83 \pm 3.22$ & $51.61 \pm 1.06$ & $49.91 \pm 0.99$ & $46.08 \pm 1.28$\\
\bottomrule
\end{tabular}
}
\end{table}
As can be seen from Table \ref{tab:noisy_label}, while those baselines achieve comparable performance at $20 \%$ noise rates, the accuracy scores of last layers tuning and linear probing drop significantly as the noise rate increases, compared to prompt tuning and CPT. In summary, CPT consistently achieves best worst-group accuracy in all settings and prompt. Interestingly, our optimization procedure helps improve prompt tuning on all metrics when the noise level is higher than $50 \%$, which shows the benefit of effectively leveraging gradient information from multiple groups in deriving the updating direction.

\begin{figure*}[!ht]
    \centering
     \includegraphics[width=.9\columnwidth, ]{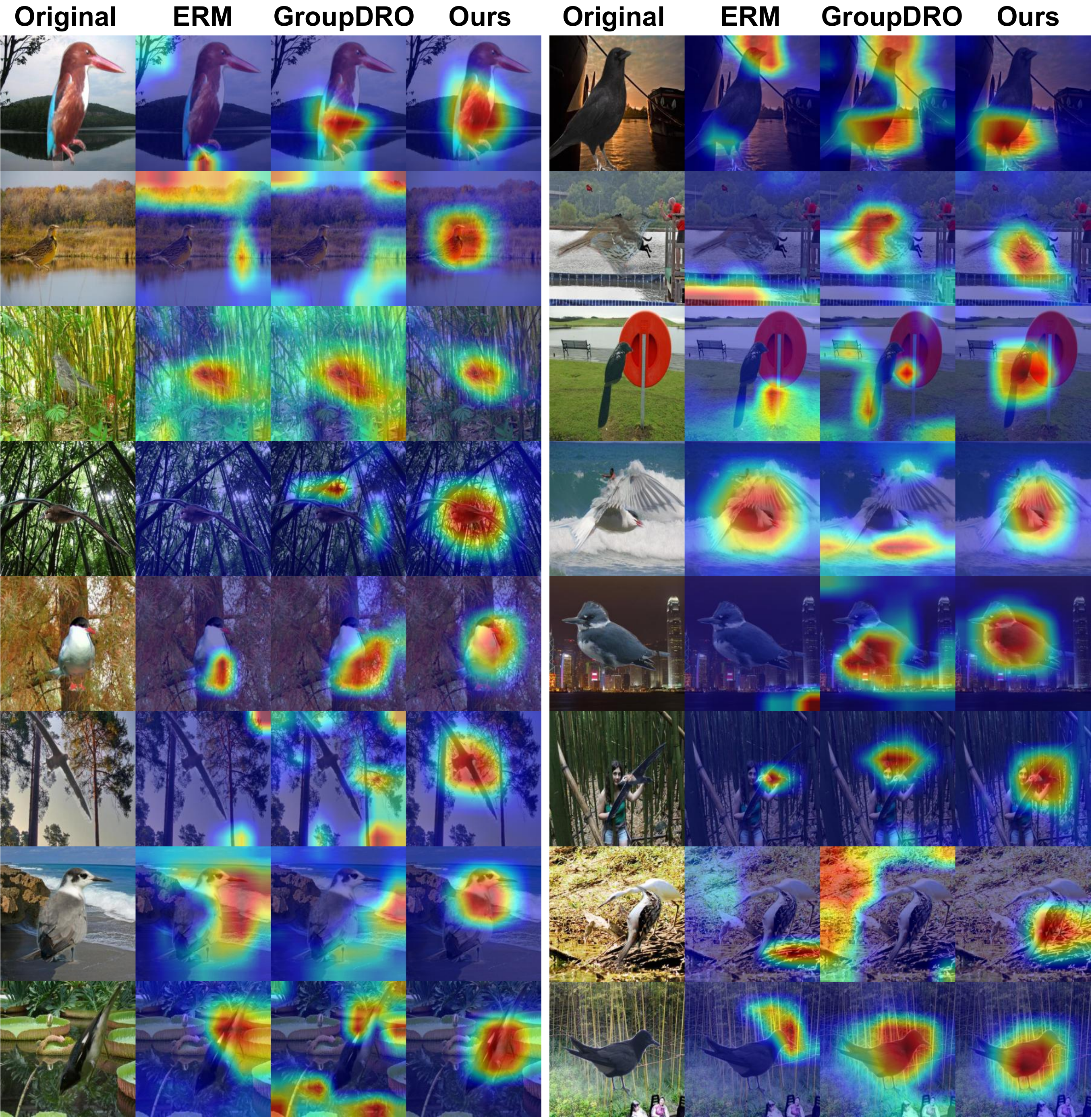}
    \caption{Waterbirds GradCAM.}
    % \vspace*{-2mm}
    \label{fig:wb_gradcam}
\end{figure*}

\clearpage
\textbf{Saliency maps.} We here provide more GradCAM visual explanations for ERM, GroupDRO and CPT on Waterbirds in Figure \ref{fig:wb_gradcam}. From those images, CPT shows its ability to identify causal features for making predictions, compared to ERM and GroupDRO. Hence, our balancing mechanism helps prevent the model from learning spurious features that occur frequently during training.

% \begin{figure*}[!ht]
%     \centering
%      \includegraphics[width=.9\columnwidth, ]{fig/celeba_gradcam.pdf}
%     \caption{CelebA GradCAM.}
%     % \vspace*{-2mm}
%     \label{fig:celeba_gradcam}
% \end{figure*}
% \clearpage
% \textbf{Prompt templates for CLIP} 

% Here, we delve into the efficiency of CPT, examining
% memory consumption and computational demands
% \begin{table}[!ht]
%   \caption{Number of trainable parameters and the storage required for all experiments [TO BE UPDATED]:}
%   \label{tab:storage}
%   \centering
%   \resizebox{0.5\columnwidth}{!}{%
% \begin{tabular}{lccc}
% \toprule
% &  \# Params  & Size & Avg. per task \\
% \hline ERM &   $94048520$ & $360\,\text{MB}$  & $90\,\text{MB}$\\
% GroupDRO   &   $94048520$  & $360\,\text{MB}$  & $90\,\text{MB}$ \\

% \midrule Ours (ResNet) &    $94048520$  & $360\,\text{MB}$  & $90\,\text{MB}$ \\
% Ours  &   $328712$ & $1.38\,\text{MB}$ & $0.34\,\text{MB}$\\
% \bottomrule
% \end{tabular}}
% \end{table}

\end{document}